\pdfoutput=1

\documentclass[11pt]{article}

\usepackage[preprint]{acl}

\usepackage{times}
\usepackage{latexsym}

\usepackage[T1]{fontenc}

\usepackage[utf8]{inputenc}

\usepackage{microtype}

\usepackage{inconsolata}

\usepackage{graphicx}

\usepackage{subfigure}
\usepackage{amsmath} 
\usepackage{tcolorbox}
\usepackage{amsfonts}  
\usepackage{booktabs}  
\usepackage{url}     
\usepackage{hyperref}   
\usepackage{enumitem}
\usepackage{microtype} 
\usepackage{nicefrac} 
\usepackage{subcaption}
\usepackage{fvextra}
\usepackage{adjustbox}
\newcounter{mytcb}
\tcbuselibrary{breakable,skins}

\usepackage{supertabular}
\usepackage{longtable}
\usepackage{array}
\usepackage{tabularx}

\newenvironment{featureset}[1]{%
  \begin{table*}
  \caption{#1}\label{tab:#1}
  \begin{supertabular}{@{}p{0.15\textwidth}p{0.85\textwidth}@{}}
  \toprule
  \textbf{Feature} & \textbf{Explanation} \\
  \midrule
}{%
  \bottomrule
  \end{supertabular}
  \end{table*}
}

\newcommand{\featurerow}[2]{%
  #1 & \begin{minipage}[t]{\linewidth}
    \small\ttfamily
    #2
  \end{minipage} \\
  \midrule
}

\title{Decoding Dark Matter: Specialized Sparse Autoencoders for Interpreting Rare Concepts in Foundation Models}

\author{
  Aashiq Muhamed\textsuperscript{1}, 
  \textbf{Mona Diab\textsuperscript{1}, 
  Virginia Smith\textsuperscript{2}} \\
  \{amuhamed, mdiab, smithv\}@andrew.cmu.edu
  \\
   \textsuperscript{1} Language Technologies Institute,
  \textsuperscript{2} Machine Learning Department \\
  Carnegie Mellon University
  }

\begin{document}
\maketitle
\begin{abstract}

\looseness=-1
Understanding and mitigating the potential risks associated with foundation models (FMs) hinges on developing effective interpretability methods.  
 Sparse Autoencoders (SAEs) have emerged as a promising tool for disentangling FM representations, but they struggle to capture rare, yet crucial concepts in the data. We introduce Specialized Sparse Autoencoders (SSAEs), designed to illuminate these elusive \emph{dark matter} features by focusing on specific subdomains. We present a practical recipe for training SSAEs, demonstrating the efficacy of dense retrieval for data selection and the benefits of Tilted Empirical Risk Minimization as a training objective to improve concept recall. Our evaluation of SSAEs on standard metrics, such as downstream perplexity and $L_0$ sparsity, show that they effectively capture subdomain tail concepts, exceeding the capabilities of general-purpose SAEs. We showcase the practical utility of SSAEs in a case study on the Bias in Bios dataset, where SSAEs achieve a 12.5\% increase in worst-group classification accuracy when applied to remove spurious gender information. SSAEs provide a powerful new lens for peering into the inner workings of FMs in subdomains.

\end{abstract}

\section{Introduction}

\looseness=-1
Interpretability is crucial for ensuring the safety and reliability of foundation models (FMs) \citep{bommasani2021opportunities}. 
A key challenge in interpretability research is to scalably explain the myriad unanticipated behaviors in FMs.
Sparse Autoencoders (SAEs) have recently emerged as a promising tool for disentangling the complex, high-dimensional representations within FMs into meaningful, human-interpretable features without supervision \citep{cunningham2023sparse, gao2024scaling, braun2024identifying, bricken2023monosemanticity}. However, even massively wide SAEs, trained on vast amounts of data, may only capture a fraction of the concepts embedded within these models~\citep{templeton2024scaling}. A significant portion of rare or highly specific concepts remain essentially invisible due to their infrequent activation. These elusive features, akin to \emph{dark matter} in the universe of interpretability, pose a significant challenge for understanding and mitigating potential risks associated with FMs. While larger SAEs did exhibit some features for rarer concepts, \citet{templeton2024scaling} found compelling evidence suggesting a vast amount of \emph{dark matter} features were still being missed. For example, they found features for some of San Francisco's neighborhoods, but their model still lacked features for smaller entities like coffee shops or street intersections. They observed that if a concept is present only once every billion tokens, we may need a billion-feature SAE to capture it reliably. This raises a critical question: can we develop more efficient methods than simply scaling SAE width to capture the tail concepts we are interested in?

\looseness=-1
This paper introduces Specialized Sparse Autoencoders (SSAEs), a novel approach designed to address this challenge. Instead of aiming to capture all concepts, as in current SAE practices, we propose SSAEs as an unsupervised targeted method for efficiently extracting rare features related to specific subdomains. By focusing on a particular subdomain, we can train SSAEs to learn features representing tail concepts without needing to scale to billions of features. Furthermore, instead of relying solely on scaling, we investigate whether Tilted Empirical Risk Minimization (TERM), which approximates minimax risk at large tilt parameters, can further improve the representation of tail concepts within SSAEs.
Our key contributions are:
\begin{enumerate} [itemsep=2pt, topsep=0pt, leftmargin=6pt, rightmargin=1pt, labelsep=3pt, itemindent=5pt, parsep=2pt]
    \looseness=-1
    \item \textbf{Specialized Sparse Autoencoders:} 
        An unsupervised method for efficiently extracting rare, subdomain-specific features. We demonstrate empirically that SSAEs capture a greater proportion of tail concepts than standard SAEs trained on general-purpose data, achieving a 12.5\% increase in worst-group classification accuracy on the Bias in Bios dataset when used to remove spurious gender information.

    \item \textbf{Subdomain Data Selection Strategies:} 
         A practical recipe for training SSAEs, starting with a small seed dataset and leveraging various data selection strategies to identify relevant training data from the FM's pretraining corpus. We find that Dense retrieval is particularly effective while TracIn reranking can offer further improvements.

     \looseness=-1
    \item \textbf{Tilted Empirical Risk Minimization for SAEs:} 
    A novel training objective for SAEs designed to improve concept recall. At large tilt values, TERM encourages more balanced learning of head and tail concepts. We show that TERM-trained SSAEs are more interpretable, exhibit improved concept detection, while maintaining comparable downstream perplexity.
    
\end{enumerate}

 \looseness=-1
 We envision SSAEs as versatile tools for concept detection and control across domains where identifying rare features is crucial, such as AI safety (detecting deception), healthcare (identifying outliers), and fairness (recognizing underrepresented groups). See \autoref{sec:applications} for additional examples.

\looseness=-1
\paragraph{Related Work}
Much interpretability research focuses on analyzing coarse-grained model components like induction heads and MLP modules \citep{olsson2022context, elhage2022toy, geva2023dissecting, meng2022locating, nanda2023fact}, or fine-grained units like linear probes \citep{kim2018interpretability, belinkov2022probing, geiger2023causal, zou2023representation}. Both have limitations. The inherent polysemanticity of coarse-grained components complicates interpretation. 
Fine-grained analysis, while potentially more precise, is constrained by reliance on curated datasets that isolate behavior, limiting generalizability to unknown mechanisms.
Feature disentanglement methods, such as SAEs \citep{bricken2023monosemanticity, cunningham2023sparse}, offer a promising unsupervised alternative, aiming to identify human-interpretable directions in an FM's latent space.  For additional work see \autoref{sec:related-work}.

\section{Methodology}

\subsection{Sparse Autoencoders (SAE)}
\looseness=-1
The superposition hypothesis in FMs suggests that a limited number of neurons encode a much larger number of concepts, leading to complex and overlapping representations \citep{elhage2022toy}. Superposition, while efficient, makes it challenging to interpret individual neuron representations or directions in representation space. 
Sparse autoencoders (SAEs) offer a potential solution by learning to reconstruct FM representations at a layer using a sparse set of features in a higher-dimensional space, potentially disentangling superposed features and revealing more interpretable representations \citep{elhage2022solu, olshausen1997sparse}. In a well-trained SAE, individual features in the hidden dimension align with underlying sparse, semantically meaningful features \citep{donoho2006compressed}.

SAEs decompose a model's activation $x \in \mathbb{R}^n$ into a sparse, linear combination of feature directions:
$
x \approx x_0 + \sum_{i=1}^{M} f_i(x) d_i, 
$
where $d_i$ are $M \gg n$ latent unit-norm feature directions, and the sparse coefficients $f_i(x) \geq 0$ are the corresponding feature activations for $x$. The right-hand side of this equation has the structure of an autoencoder: an input activation $x$ is encoded into a (sparse) feature activations vector $f(x) \in \mathbb{R}^M$, which is then linearly decoded to reconstruct $x$.
We parameterize a single-layer autoencoder $(f, \hat{x})$ as follows:
$f(x) := \text{ReLU}(W_{enc}(x) + b_{enc}) $ and 
$\hat{x}(f) := W_{dec}f + b_{dec}$
where $W_{enc} \in \mathbb{R}^{M \times n}$ and $W_{dec} \in \mathbb{R}^{n \times M}$ are the encoding and decoding weight matrices, and $b_{enc} \in \mathbb{R}^M$ and $b_{dec} \in \mathbb{R}^n$ are the bias vectors.
The training objective combines a reconstruction loss and a sparsity penalty:
\begin{equation}
    \label{eq:train}
    L(x) = \|x-\hat{x}(f(x))\|_2^2+\lambda\|f(x)\|_1
\end{equation}

where $\lambda > 0$ is a hyperparameter controlling the trade-off between reconstruction fidelity and sparsity. We constrain the columns of $W_{dec}$ to have unit norm during training  \citep{bricken2023monosemanticity}. 

In existing work, SAEs for FMs are trained on the same large, general-purpose dataset used to train the underlying FM \citep{bricken2023monosemanticity, cunningham2023sparse, rajamanoharan2024improving, gao2024scaling}. This approach ensures that the SAE captures a wide array of concepts present in the general language domain. However, this can result in the SAE learning features that are frequent in the pretraining data but miss concepts within specific domains of interest, especially those that are rare by frequency in the pretraining data.

\subsection{Specialized Sparse Autoencoders (SSAE)}
\looseness=-1
Specialized Sparse Autoencoders are designed to learn features representing rare concepts within specific subdomains. Our approach begins with a small seed concept dataset, comprising either a specific concept or limited data from the target subdomain (e.g., toxicity). We then expand this seed dataset using a high-recall retrieval strategy that leverages the seed data to identify and retrieve subdomain-relevant examples from the base FM's pretraining corpus. 
To create an SSAE, we finetune a pretrained general-purpose SAE (GSAE) on this curated subdomain data using \autoref{eq:train}. The GSAE is initially trained to reconstruct activations on a large, general-purpose dataset, enabling it to capture a broad range of concepts. Finetuning on the subdomain data allows the SAE to specialize and learn features that may be infrequent in the general domain but prevalent within the target subdomain.

To evaluate the quality of the trained SAEs, we use $L_0$ and Perplexity with SAE \citep{bricken2023monosemanticity}.
\textbf{$L_0$} measures the sparsity of the SAE and is defined as the average number of active features on a given input, i.e. $\mathbb{E}_{x \sim D} \| f(x) \|_0$.
Perplexity with SAE measures the reconstruction fidelity of the SAE and is the average cross-entropy loss of the language model on an evaluation dataset, when the SAE's reconstructions are spliced into it. A better SAE recovers more of the base model's performance. All other things being equal, a better SAE needs fewer features ($L_0$) to explain model performance on a given datapoint. Unlike existing works that evaluate SAEs on subsampled training data, we evaluate SSAE generalization using both in-distribution and out-of-distribution test sets drawn from the same subdomain. This dual evaluation approach assesses the SSAE's ability to both accurately capture concepts within the specific training data distribution and generalize to unseen data, reflecting the capability to learn broader subdomain concepts.
Additionally, we perform automated interpretability scoring and qualitative analysis, to verify the interpretability of the learned features.

\subsection{Subdomain Data Selection Strategies}
\looseness=-1
 SSAE effectiveness depends on the quality and relevance of the selected subdomain data used for finetuning. We study several selection strategies to identify data points from a larger corpus (FM's pretraining data) most relevant to the seed data:

\paragraph{Sparse Retrieval:}
\looseness=-1
Okapi BM25 \citep{robertson2009probabilistic}, a TF-IDF variant, ranks documents based on query relevance, considering term frequency, inverse document frequency, and document length. We use the seed dataset as query to retrieve relevant documents from the larger corpus.

\paragraph{Dense Retrieval:}
\looseness=-1
Contriever \citep{izacard2022unsupervised}, a dual-encoder dense retriever, generates semantically meaningful embeddings for queries and documents. We embed the seed dataset and candidate documents, using cosine similarity to retrieve documents most similar to the seed concepts.

\paragraph{SAE TracIn:}
Training data Influence Score (TracIn) \citep{pruthi2020estimating} quantifies training examples' influence on model predictions. We adapt TracIn to SAEs by calculating the dot product of the loss gradients with respect to the training data and seed data: 
$
    \operatorname{TracIn}\left(z, z^{\prime}\right)=\nabla L_w\left(z\right) \cdot \nabla L_w\left(z^{\prime}\right)
$
where $z$ is a training data point, $z'$ is the seed dataset, $w$ are the pretrained SAE weights, and $L_w(\cdot)$ is the SAE loss (\autoref{eq:train}). 
We use a two-stage approach to identify influential data: Initial Filtering with Sparse/Dense retrieval, then TracIn Reranking to select points for SSAE training.
\subsection{Tilted Empirical Risk Minimization for Enhanced Detection}

\looseness=-1
Finetuning with Empirical Risk Minimization (ERM) tends to prioritize learning features for the most frequent \emph{head} concepts in the subdomain data.
However, for many applications such as safety, capturing rare \emph{tail} concepts is often crucial. 
These rare features may represent potential risks or safety violations and are often overlooked by standard ERM as it focuses on minimizing the average loss. The objective then is not minimizing the average loss, but rather minimizing the maximum risk to ensure that even the rarest, potentially dangerous features are captured.  Tilted Empirical Risk Minimization (TERM) \citep{li2020tilted, beirami2018characterization} provides a framework for approximating this minimax risk, encouraging the model to learn features that better represent these tail concepts.

\looseness=-1
TERM modifies the standard ERM objective by introducing a tilt parameter ($t$) that controls the emphasis on different parts of the loss distribution:
$
\tilde{L}(t ; w)=\frac{1}{t} \log \left(\frac{1}{N} \sum_{i \in[N]} e^{t \cdot L_w\left( z_{i}\right)}\right)
$
where $L_w(z_i)$ is the standard SAE loss (\autoref{eq:train}) for data point $z_i$ in a minibatch with $N$ points and SAE parameters $w$.
TERM generalizes ERM as the 0-tilted loss recovers the average loss, while it also recovers other alternatives such as the max-loss ($t \rightarrow +\infty$) and min-loss ($t \rightarrow -\infty$). In this work we use large tilt parameters ($t \gg 0$) to effectively minimize the maximum loss, encouraging the model to learn features that better represent the tail of the data distribution, including rare concepts.
Minimax losses are also known to enhance robustness to OOD data, which is relevant for detecting rare concepts often underrepresented in training data \citep{ye2021towards, sagawa2019distributionally}.

\begin{figure*}[htb]
    \centering
    \includegraphics[width=0.45\textwidth]{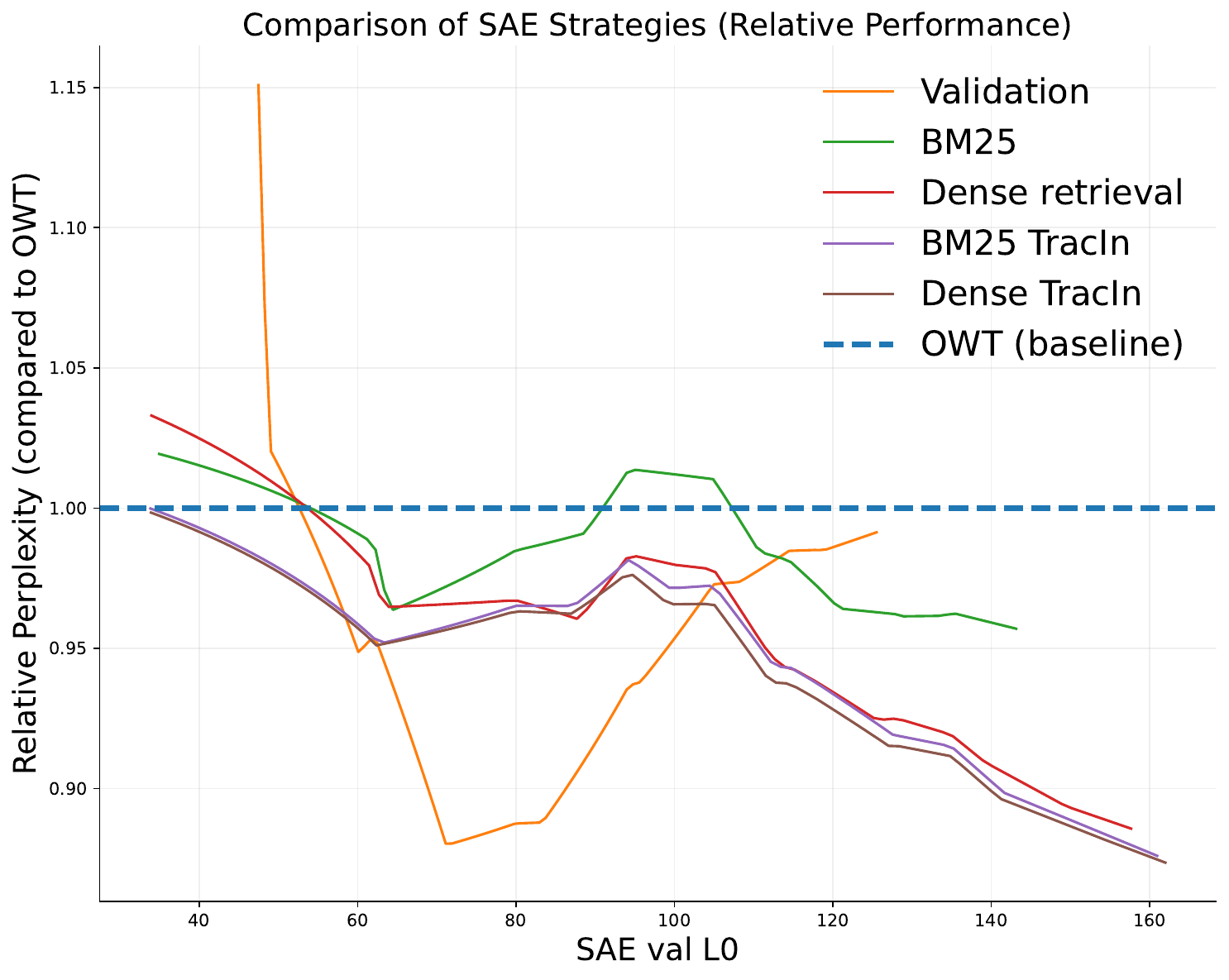}
    \hfill
    \includegraphics[width=0.45\textwidth]{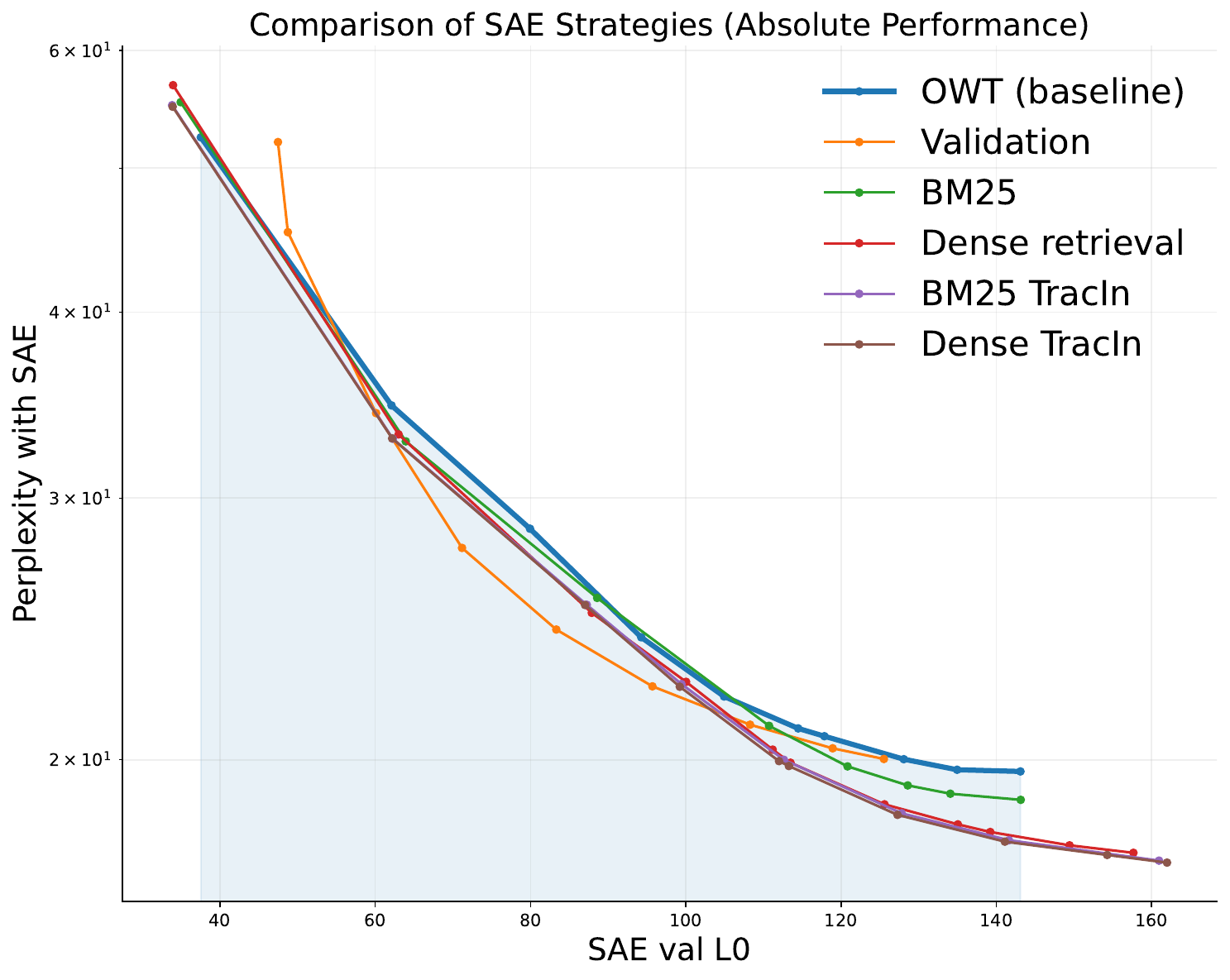}
    \caption{\small Pareto curves for Physics SSAE trained with various data selection strategies as the $\lambda$ is varied on arXiv Physics test data. We plot (Left) Perplexity with spliced in SSAE relative to GSAE baseline and (Right) Absolute Perplexity with spliced in SSAE. Dense TracIn and BM25 TracIn achieve comparable performance, performing slightly better than Dense retrieval, which outperforms BM25 retrieval and OWT Baseline. All curves are averaged over three SAE training seeds.}
    \label{fig:physics}
\end{figure*}

\looseness=-1
Incorporating TERM during finetuning leads to a more balanced representation of both head and tail concepts within the subdomain. This shift reflects a fundamental trade-off between precision and recall in TERM-trained and ERM-trained SAEs. Standard ERM prioritizes precision, yielding highly specialized features that allow for fine-grained control over concepts but may miss rare ones. TERM prioritizes recall, sacrificing some control for broader concept coverage, particularly of rare concepts, making it advantageous for detecting potentially harmful behaviors. TERM encourages the SAE to learn compositional features leading to more interpretable representations (see \autoref{sec:proof} for a formal argument).

\section{Experiments And Results}
\subsection{Specialized Sparse Autoencoders (SSAEs)}
\subsubsection{Data Selection Strategies}
\looseness=-1
In this section, we evaluate the effectiveness of various data selection strategies for training SSAEs.
\paragraph{Experimental Setup}
\label{sec:exp-setup}

\looseness=-1
We use the pretrained Gemma-2b \citep{team2024gemma} residual stream GSAE 
({gemma-2b-res-jb} checkpoint at {blocks.12.hook\_resid\_post} layer) \citep{huggingfaceGemma}.
These SAEs have feature width 16384 and were pretrained on OpenWebText (OWT) \citep{Gokaslan2019OpenWeb}. 
For the Pareto front, we sweep 8 L1 penalty coefficients, selecting the best model on validation for each L1 value, then evaluating on the held-out test split.
SAEs are trained using Adam \citep{kingma2015adam} with lr 5e-5, token batch size 4096, data shuffled within a batch buffer of size 4, and linear lr decay over the last 1000 steps. Experiments complete in under 12 hours using 4 A6000 GPUs. We use SAELens \citep{bloom2024saetrainingcodebase} for training and analysis.

\paragraph{SSAE for Physics}
\looseness=-1
We start with a seed concept dataset (Validation) consisting of 9.2K tokens sampled from the arXiv Physics dataset \citep{huggingfaceArxivPhysics}.
Using BM25, Dense Retrieval, and SAE TracIn, we expand this to 13.9M tokens from OWT. The SSAE is trained by finetuning the GSAE for 1000 iterations on this expanded set.
For SAE TracIn, we first reduce OWT to 1\% using BM25 or Dense retrieval, then rerank using TracIn scores and select 13.9M tokens. We call these methods BM25 TracIn and Dense TracIn, respectively. \\
\looseness=-1
\indent We train an SSAE for each strategy and compare its performance to a baseline SAE finetuned on the full OWT dataset across various sparsity coefficients ($\lambda$). We evaluate the models on two test splits: 4.8M tokens from arXiv Physics (in-distribution) and 700K tokens from Physics instruction tuning \citep{huggingfaceArxivPhysicsInstruct}(out-of-distribution). Testing on instruction data helps measure whether the SAEs are overfitting to the specific template of the text as opposed to identifying concepts. Figure \ref{fig:physics} and \ref{fig:physicsood} show the patched perplexity vs. $L_0$ curves for these experiments. \\
\looseness=-1
\indent We measure performance using area under the curve for a range of $L_0$ from 60 to 140 i.e., a selection strategy with lower perplexity (SSAE spliced in) is better. Our findings show Dense TracIn and BM25 TracIn achieve comparable performance, surpassing Dense retrieval alone, which in turn outperforms BM25 retrieval. Training on the full OWT dataset yields the lowest performance. We observe: (a)
Dense retrieval consistently outperforms BM25. SSAEs trained with Dense Retrieval achieve lower perplexity for a given $L_0$ than those with BM25, both in and out of distribution.
(b) BM25 exhibits poor out-of-distribution generalization. While BM25 performs reasonably well in-distribution, its performance degrades significantly on the out-of-distribution test set.
(c) Multiple passes on seed data (Validation) during SSAE training improve in-distribution performance but degrade out-of-distribution performance. This suggests multiple passes can overfit to the structure or template of the seed dataset.
(d) While TracIn reranking after Dense retrieval yields a marginal performance gain, Dense retrieval alone remains highly competitive.

\paragraph{SSAE for Toxicity}
We repeat the experiment on the Pile Toxicity dataset \citep{huggingfacePile} in \autoref{sec:dataselectiontoxicity}. The results align with the physics experiment, with Dense retrieval outperforming BM25 and TracIn offering a marginal improvement over Dense retrieval alone.

\subsubsection{Probing Tail Concept Learning}

\begin{figure}[t]
    \centering
    \includegraphics[width=0.5\textwidth]{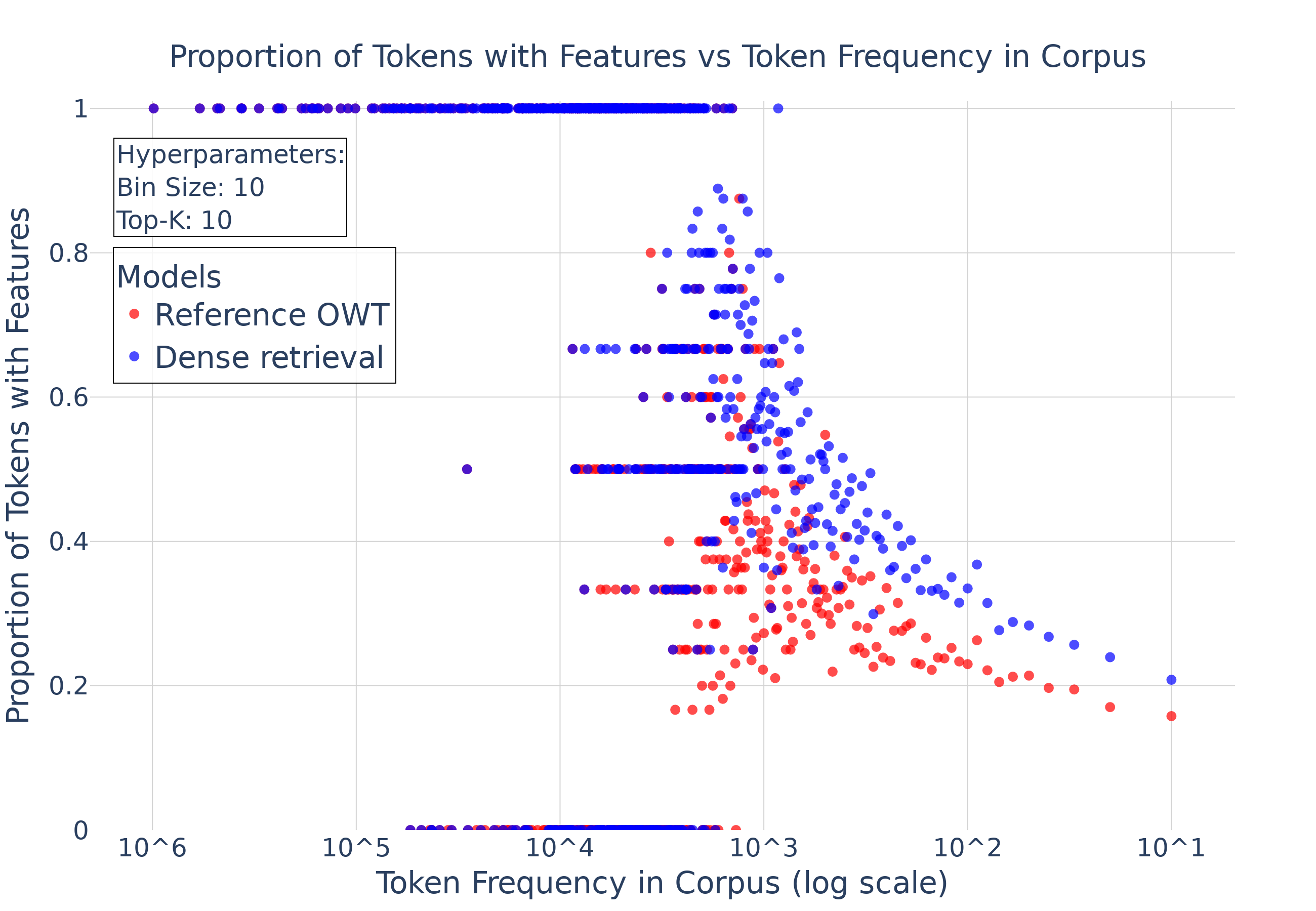}
    \caption{\small Proportion of tokens with SAE features vs. Token frequency in Physics arXiv data. SSAE trained with dense retrieval captures more tail tokens (concepts) in its features.}
    \vspace{-0.2in}
    \label{fig:tailA}
\end{figure}

To probe tail concept learning we use convergent validity \citep{campbell1959convergent} with the Logit Lens \citep{bloom2024understandingfeatureslogitlens}. Figure \ref{fig:tailA}, uses the unembedding matrix as a logit lens to analyze the top-10 token logits associated with each SSAE feature. For each frequency bucket in the Physics arXiv test data, we calculate the percentage of tokens that appear among the top-10 logits for at least one feature. This measures the extent to which SSAE features represent tokens across different frequency ranges.
We compare two SSAEs at test $L_0$ of 100: one finetuned on full OWT dataset, another using Dense retrieval. The Dense retrieval finetuned SSAE captures a significantly higher proportion of tail tokens in its features compared to the OWT finetuned SSAE. Moreover, these captured tail tokens often correspond to physics-specific concepts, suggesting that SSAEs are indeed learning to represent rare, domain-relevant concepts. Similar results are obtained for toxicity data in \autoref{fig:tailtoxicA}.

\subsubsection{Case study: Removing Spurious Features in Bias in Bios Classifier}

\looseness=-1
Having shown the effectiveness of ERM-trained SSAEs in capturing tail concepts for finer control, we apply them to the Spurious Human-interpretable Feature Trimming (SHIFT) method \citep{marks2024sparse}. SHIFT addresses the issue of FM classifiers relying on unintended signals (e.g., spurious features) by modifying their generalization through feature circuit editing. Unlike approaches that rely on disambiguated labeled data, SHIFT operates even when such data is unavailable \citep{zech2018confounding, ngo2022alignment, casper2023open, hase2024unreasonable}.
We show that replacing the GSAE with our SSAE in SHIFT further enhances its editing capabilities.

\paragraph{Method.} 
\looseness=-1
SHIFT operates as follows, given labeled training data $D = {(x_i, y_i)}$, classifier $C$ trained on $D$, and SAEs for components of $C$:
\begin{enumerate} [itemsep=0pt, topsep=0pt, leftmargin=8pt, rightmargin=1pt, labelsep=2pt, itemindent=2pt, parsep=0pt]
    \item Compute a feature circuit (see \autoref{sec:detailscircuits}) explaining $C$'s accuracy on inputs $(x, y) \sim D$ (using metric $m = -\log C(y|x)$).
    \item Manually or automatically inspect and evaluate each feature's task-relevancy.
    \item  Ablate features deemed task-irrelevant to obtain a modified classifier $C'$.
    \item (Optional) Finetune (retrain) $C'$ on data from $D$ to potentially restore performance.
\end{enumerate}

\paragraph{Experimental Setup.} 
We use the Bias in Bios dataset (BiB) \citep{de2019bias} to illustrate SHIFT with SSAEs.  BiB contains professional biographies and the task is to classify an individual's profession, with gender being a spurious feature. Two subsets are created from BiB: the \emph{ambiguous set} (male professors and female nurses) and the \emph{balanced set} (equal numbers of male professors, male nurses, female professors, and female nurses) \citep{marks2024sparse}. The ambiguous set represents a worst-case scenario where the unintended signal (gender) perfectly predicts training labels (profession). Our goal is to achieve accurate profession classification on the balanced set using only the ambiguous set for training.

\looseness=-1
Our base model is a Pythia-70M linear classifier \citep{biderman2023pythia}, trained on the ambiguous set (details in \autoref{sec:detailsshift}). 
SHIFT is applied by discovering a circuit using the zero-ablation variant (\autoref{sec:detailscircuits}).
Instead of using human judgement to ablate features, we employ Feature skyline \citep{marks2024sparse}, sweeping across 1-200 circuit features most causally implicated in spurious feature accuracy on the balanced set. The number of features to ablate is chosen based on best profession classification performance on the dev set.

We use GSAEs (width 32768) for the MLP output, attention output, and residual stream for each layer, pretrained on 2B tokens (first 128 tokens of random documents) from The Pile \citep{gao2020pile}. The SSAE is trained by retrieving 8M tokens from The Pile using a dense retriever, guided by 5 BiB examples, and finetuning all the GSAEs in every layer on this data for one epoch. We use $\lambda = 0.1$ and learning rate $10^{-4}$ throughout.

We also conduct a \emph{Compression} experiment, where we slice the GSAE to width 4096 by taking only the first 4096 rows of the decoder (Comp. GSAE). This examines a worst-case scenario where the GSAE may not capture all relevant subdomain features. Comp. SSAE is initialized with Comp. GSAE before finetuning on the retrieved tokens.

In addition to the Oracle (trained on ground-truth labels from the balanced set) and Original (trained on ground-truth labels from the ambiguous set) classifiers, we include the following baselines:
\begin{itemize} [itemsep=0pt, topsep=0pt, leftmargin=10pt, rightmargin=1pt, labelsep=2pt, itemindent=2pt, parsep=0pt]
\item Concept Bottleneck Probing (CBP): Adapted from \citet{yan2023robust} (see \autoref{sec:detailscbp}).
\item Neuron skyline: Sweeps over number of neurons to ablate (1-200) and mean-ablates those most implicated in spurious feature accuracy.
\end{itemize}

\begin{table}[htbp]
\small
\centering
\footnotesize
\setlength{\tabcolsep}{4pt}
\begin{tabular}{@{}lccc@{}}
\toprule
& \multicolumn{3}{c}{Accuracy} \\
\cmidrule(l){2-4}
Method & $\uparrow$Prof. & $\downarrow$Gen. & $\uparrow$Worst \\
\midrule
Original         & 61.9 & 87.4 & 24.4 \\
CBP              & 83.3 & 60.1 & 67.7 \\
Neuron skyline   & 75.5 & 73.2 & 41.5 \\
GSAE SHIFT           & 88.5 & 54.0 & 76.0 \\
\textcolor{blue}{SSAE SHIFT}            & 90.2 & 53.4 & 88.5 \\
GSAE SHIFT+retrain         & 93.1 & 52.0 & 89.0 \\
\textcolor{blue}{SSAE SHIFT+retrain}         & \textbf{93.4} & \textbf{51.9} & \textbf{89.5} \\
\midrule
Comp. GSAE SHIFT      & 80.5 & 68.2 & 48.6 \\
\textcolor{blue}{Comp. SSAE SHIFT}     & 89.6 & 52.2 & 78.8\\
Comp. GSAE SHIFT+retrain    & 80.0 & 68.8 & 57.1\\
\textcolor{blue}{Comp. SSAE SHIFT+retrain}   & \textbf{93.2} & \textbf{52.1} & \textbf{88.5} \\
\midrule
Oracle           & 93.0 & 49.4 & 91.9 \\
\bottomrule
\end{tabular}
\caption{\footnotesize \looseness=-1 Balanced set accuracies for intended (profession) and unintended (gender) labels. \emph{Worst} refers to lowest profession accuracy among male professors, male nurses, female professors, and female nurses. Comp.: Compressed SAE (sliced to 1/8th width). Best results per method category are bolded. }
\label{tab:accuracies}
\end{table}

\paragraph{Results.} As shown in Table \ref{tab:accuracies}, GSAE SHIFT effectively reduces the classifier's dependence on gender compared to baselines such as CBP, with Step 3 (feature ablation) providing the most substantial improvement. Applying SHIFT with neurons (Neuron skyline) performs worse than SHIFT with SAEs, likely due to the polysemantic nature of individual neurons \citep{marks2024sparse}.

\looseness=-1
SHIFT with SSAEs further improves performance, achieving a $1.7\%$ increase in profession accuracy, a $12.5\%$ increase in worst-group accuracy, and a decrease in spurious gender accuracy, demonstrating its superiority to GSAEs in fine-grained control. These gains persist even after retraining the probe, albeit to a smaller extent. The improvement can be attributed to the SSAE activating more sundomain-relevant features. For instance, at an activation threshold of 0.01, the SSAE activates 908 features compared to 602 in the GSAE. These additional features, as explored in \autoref{sec:detailscircuits}, play a crucial role in the sparse feature circuits of the classifier, explaining more of the variance previously attributed to error nodes by the GSAE.

\looseness=-1
In the Compression experiment, the performance of the Comp. GSAE with missing features drops significantly compared to the GSAE. Retraining the probe fails to mitigate this performance loss, and SHIFT is ineffective at removing spurious features with Comp. GSAE. However, the Comp. SSAE recovers most of this lost performance, even surpassing GSAE SHIFT by 1.1\% in profession accuracy. Retraining the probe with Comp. SSAE restores nearly all lost performance.

\subsection{Tilted ERM for Enhanced Detection}
\subsubsection{Motivating Example: TERM-trained GSAEs on TinyStories}

ERM-trained GSAEs prioritize learning frequent concepts in the data. In this section, we study features in TERM-trained GSAEs, showing that TERM improves feature recall at the expense of feature control.

\paragraph{Experimental Setup.} We use the 8-layer, 1M parameter base model \texttt{TinyStories-1M} \citep{eldan2023tinystories}. SAEs of width 64 are trained on the residual stream of the 7th layer using both ERM and TERM (tilt=$10^9$). We use batch size 64, lr $10^{-3}$, $\lambda=0.01$, and train for 1 epoch on the \texttt{roneneldan/TinyStories} dataset. We report results on checkpoints with $L_0$ of 16.

\paragraph{Results}

\begin{figure}[htbp]
    \centering
    \includegraphics[width=0.5\textwidth]{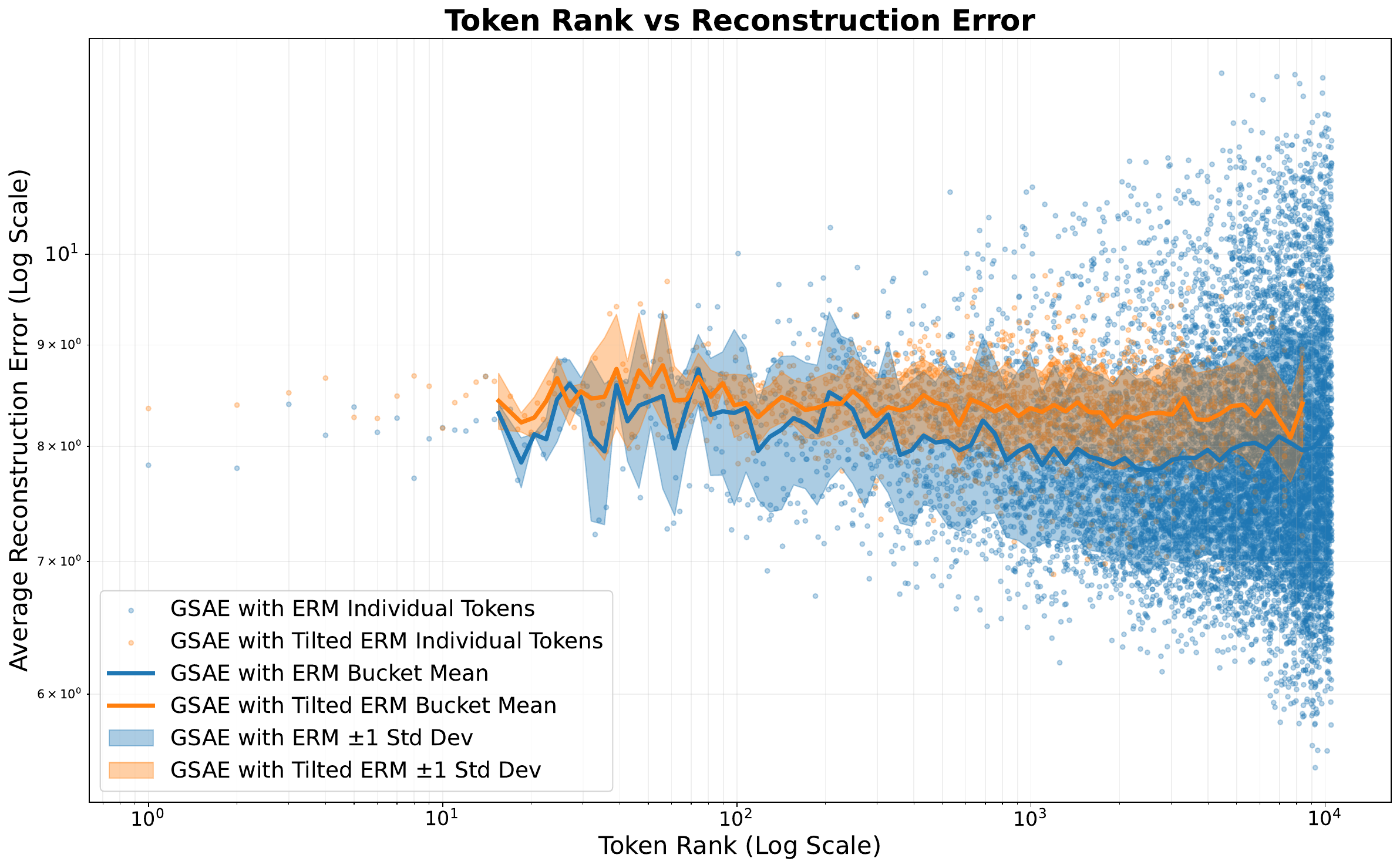}
        \caption{\small Reconstruction error vs. token rank for TERM-trained and ERM-trained GSAEs. TERM exhibits lower error variance and maximum error for tail tokens.}
    \label{fig:etokenrankerror}
\end{figure}

\looseness=-1
Figure \ref{fig:errorhistogram} shows the distribution of reconstruction error for the TERM-trained GSAE. TERM minimizes max error at the cost of slightly higher average error. Figure \ref{fig:etokenrankerror} plots the reconstruction error for tokens ranked by frequency, showing that TERM reduces reconstruction error and error variance for tail tokens compared to ERM.

\looseness=-1
We analyze decoder feature vector coverage using three approaches. Figure \ref{fig:umapplot} presents a UMAP visualization of token activations and decoder features for both GSAEs, revealing a greater dispersion of decoder directions for the TERM-trained GSAE, indicating broader coverage. Figure \ref{fig:decodercoverage} quantifies the distribution of cosine similarities between decoder directions, with the TERM-trained GSAE showing lower overall similarity, suggesting greater coverage. Figure \ref{fig:pcacoverage} shows that the TERM-trained GSAE requires more PCA components to explain variance in decoder feature directions (40) compared to the ERM-trained GSAE (21). Taken together, this shows that TERM-trained SAEs cover a wider range of features than ERM-trained SAEs.

\begin{figure}[htbp]
    \centering
    \includegraphics[width=0.5\textwidth]{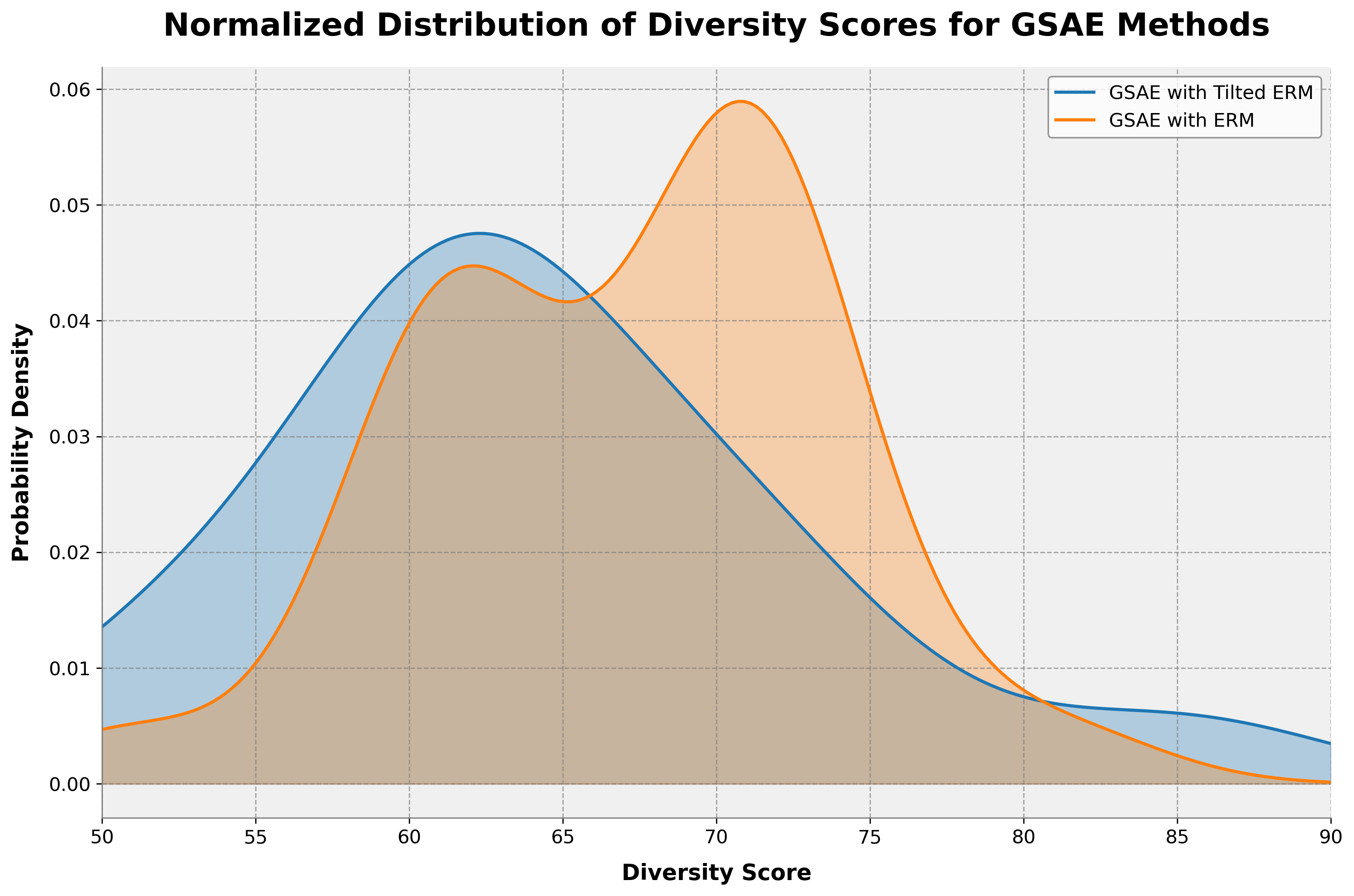}
    \caption{\small Feature diversity score distributions for TERM-trained and ERM-trained GSAEs. TERM leads to both higher and lower diversity features. Lower diversity features specialize in tail concepts, while higher diversity features capture a broader range of concepts.}
    \label{fig:gsaedist}
\end{figure}

Figure \ref{fig:gsaedist} presents diversity score distributions for TERM- and ERM-trained GSAE feature explanations (examples in \autoref{sec:tinystories-features}), capturing the variety of examples explainable by each feature using Claude 3.5 Sonnet (see Section \ref{sec:feature-diversity}). TERM-trained GSAEs exhibit both higher and lower diversity features compared to ERM, with lower diversity features specializing in tail concepts and higher diversity features capturing a broader range of concepts, both frequent and rare.

\begin{figure}[htb]
    \centering
    \includegraphics[width=0.48\textwidth]{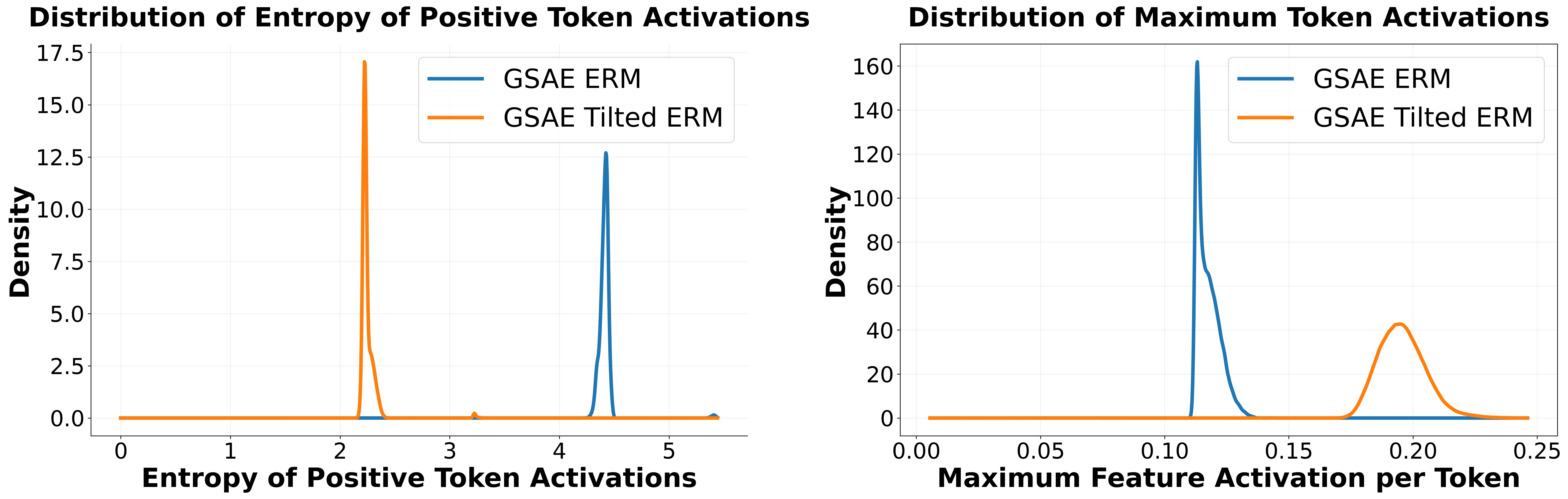}
        \caption{\small TERM feature activation patterns. (Left) TERM token activation entropy is lower, suggesting more specialized features. (Right) TERM max feature activations per token are higher. These characteristics, from minimizing max risk, contribute to TERM's enhanced tail concept detection.}
    \label{fig:tokenentropy}
\end{figure}

\looseness=-1
Figures  \ref{fig:tokenentropy} and \ref{fig:tokensthreshold} show that TERM-trained GSAE features exhibit stronger activations and lower entropy compared to ERM-trained GSAE on the data. 
This, combined with their high recall, suggests a strategy for rare concept detection: tag features strongly associated with rare concepts during pretraining, and at test time, strong activation of these tagged features triggers further investigation.
This is more effective than using error nodes with ERM-trained SAEs for rare concept detection, as error nodes do not disambiguate types of rare features.

\subsubsection{TERM-trained SSAE Performance}

\looseness=-1
While ERM-trained SSAEs improve tail concept coverage compared to GSAEs, they still prioritize learning frequent subdomain concepts. 
TERM-trained GSAEs could potentially offer better tail concept representation, but training SAEs from scratch is computationally expensive \citep{lieberum2024gemma}. Therefore, we investigate whether finetuning SSAEs with TERM on the retrieved data (using hyperparameters from Sec \ref{sec:exp-setup}) can achieve similar properties to TERM-trained GSAEs.

\begin{figure}[htb]
    \centering
    \includegraphics[width=0.5\textwidth]{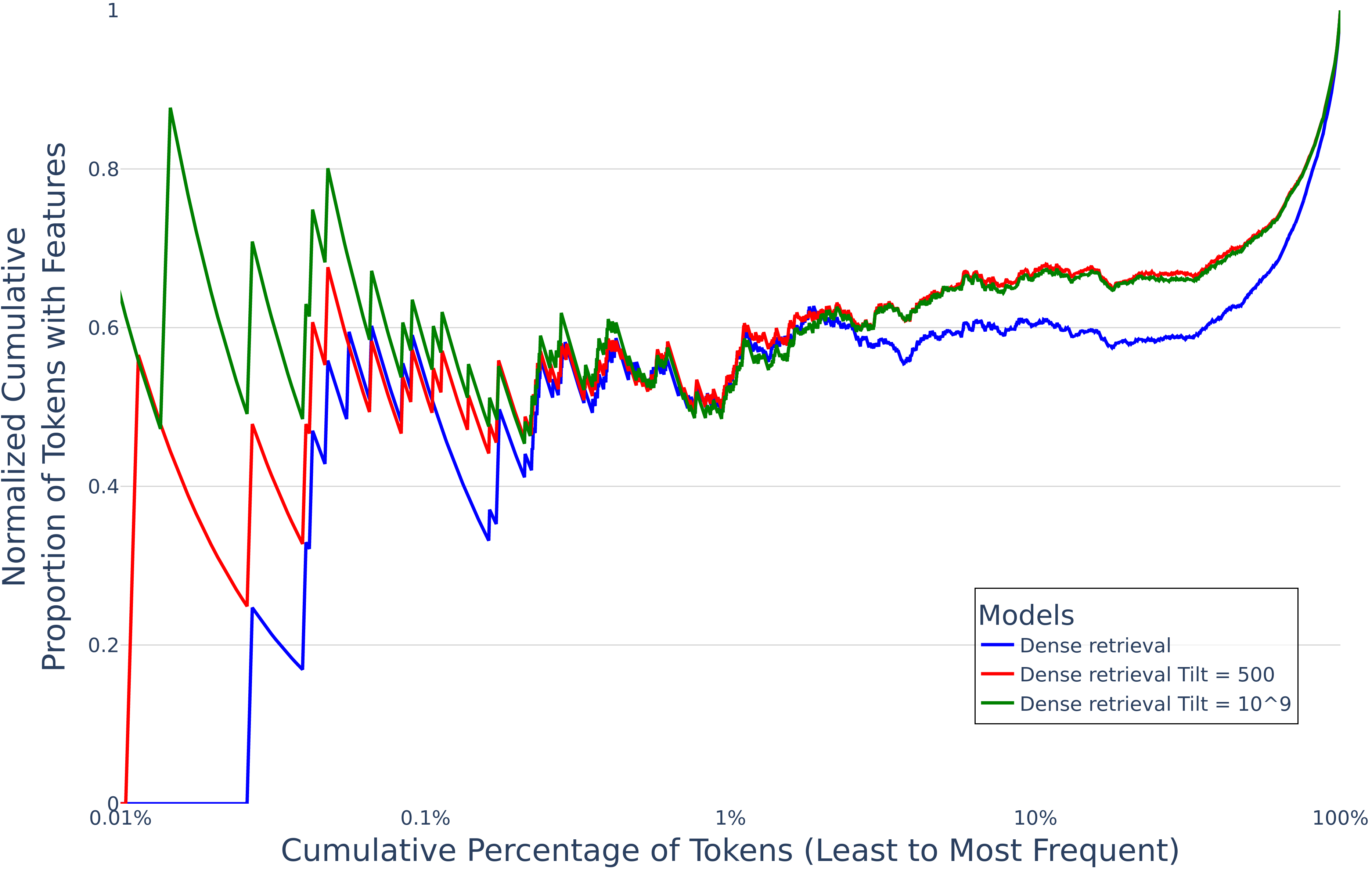}
    \caption{\small Cumulative proportion of tokens with SAE features vs. cumulative percentage of tokens in Physics arXiv data, normalized per model so that the cumulative proportion of tokens with features is 1 over the entire dataset. SSAE trained with dense retrieval and larger tilt captures more tail tokens (concepts) in its features.}
    \label{fig:tailB}
\end{figure}

\looseness=-1
\paragraph{Enhanced Tail Concept Capture with TERM}
Figure \ref{fig:tailB} plots the cumulative proportion of tokens with SSAE features (identified using the logit lens approach) versus the cumulative percentage of tokens in the Physics arXiv data for different SSAEs. We normalize the curves per model at a validation $L_0$ of 100, so that the cumulative proportion of tokens with features is 1 over the entire dataset. Results show that SSAEs trained with Dense retrieval and tilt capture a greater proportion of tail tokens compared to Dense retrieval alone, with this effect increasing with tilt. Figure \ref{fig:tailtoxicB} shows a similar trend for the Toxicity dataset.

\begin{figure}[htb]
    \centering
    \includegraphics[width=0.48\textwidth]{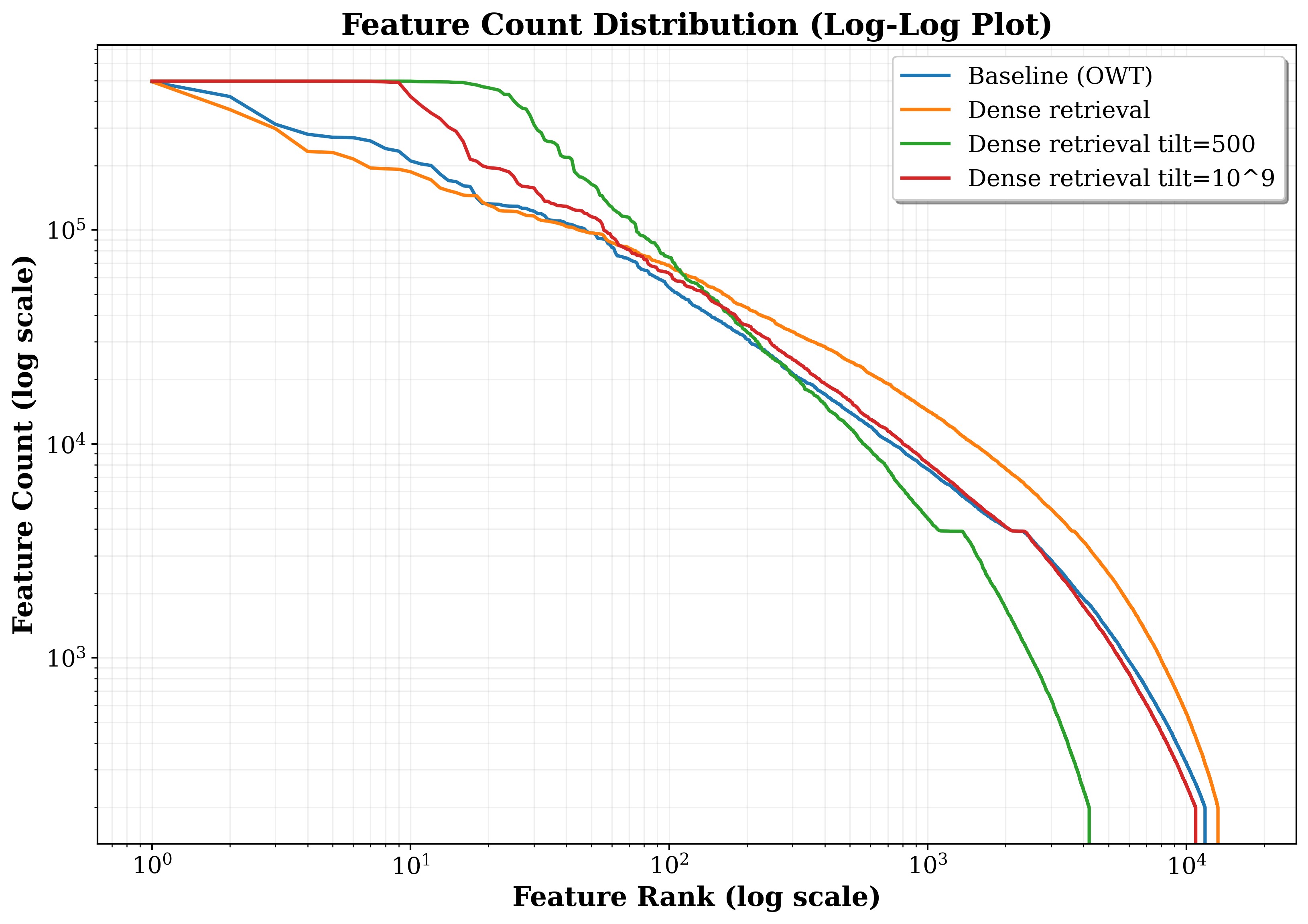}
    \caption{\small \textbf{Feature activation count vs. feature rank} for SSAEs trained on the Physics arXiv dataset using different strategies: full OWT, Dense retrieval, and Dense retrieval with tilt. Tilt encourages the learning of more broadly activating features, indicating increased concept coverage and recall.}
    \label{fig:activation_rank}
\end{figure}

\paragraph{Feature activation counts} 
\looseness=-1
Figures \ref{fig:relativehistogram2} and \ref{fig:relativehistogram3} analyze the distribution of differences in feature activation counts between SSAEs (both ERM and TERM-trained) and the OWT baseline on the Physics arXiv test set. 
The peak at 0 indicates that SSAEs retain some similarity to the baseline in their activation patterns. 
The ERM-trained SSAE exhibits greater probability mass on the right, indicating a focus on frequent concepts, while the TERM-trained SSAEs shift probability mass leftward as tilt increases, suggesting a stronger emphasis on representing domain-specific tail concepts.

\looseness=-1
Figure \ref{fig:activation_rank} plots feature activation count vs. feature rank, showing that TERM with large tilt encourages learning more broadly activating features with increased concept recall. 
This represents a fundamentally different mechanism for feature learning compared to standard ERM, promoting more compositional features that capture tail concepts.

\looseness=-1

\looseness=-1
\paragraph{Downstream perplexity}

Figures \ref{fig:tilteval} and \ref{fig:tiltoxicteval} show that TERM-finetuned SSAEs achieve comparable downstream perplexity to ERM-finetuned SSAEs within the typical $L_0$ regime used.
However at very large or low $L_0$, training with Adam can lead to higher average risk or many inactive features. Adaptive penalty schemes offer a promising solution to this challenge (see \autoref{sec:tiltpareto}).

\subsubsection{Automated Interpretability}

\begin{figure}[htbp]
    \centering
    \includegraphics[width=0.48\textwidth]{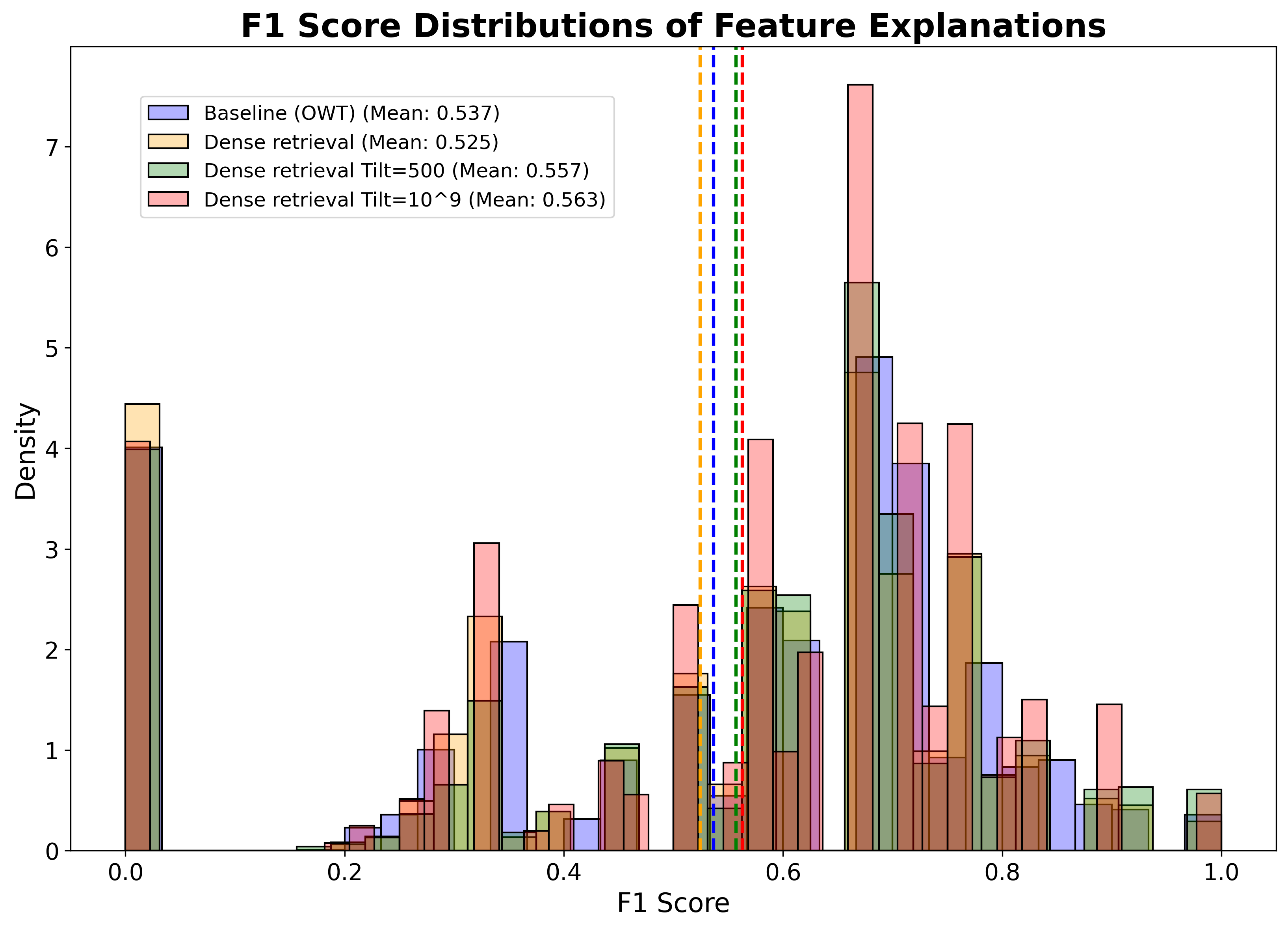}
    \caption{\small \textbf{Automated interpretability}: 
    F1 score distributions for predicting feature activation on Physics arXiv, using only FM-generated explanations. An LM is given examples activating a feature and asked to generate an explanation, which is then used to predict activations on new examples. Dense retrieval with tilt produces more predictive explanations than both the OWT baseline and Dense retrieval alone.
    }
    \label{fig:interpretability}
\end{figure}

\looseness=-1
We employ a sequence-level classification task to evaluate interpretability \citep{bills2023language, templeton2024scaling}. Instead of predicting feature activation at each token, an FM is tasked with identifying whether entire sequences contain a given feature. This simplifies the task, producing reliable scores even with smaller, faster FMs \citep{juang2024autointerp}.
Using Claude 3.5 Sonnet \citep{claude3.5sonnet} as both the \emph{Interpreter} and the \emph{Predictor}, our framework tasks the Interpreter with generating explanations for each feature based on the top 10 activating examples (see Appendix \ref{sec:explanations}). The Predictor then receives these explanations along with 10 examples (5 activating, 5 non-activating) and predicts whether each example activates the feature (see Appendix \ref{sec:autointerp_prompts} for prompts). We measure explanation interpretability using the F1 score between the Predictor's predictions and the true feature activations.

\looseness=-1
Fig \ref{fig:interpretability} shows that TERM-trained SSAEs achieve higher F1 scores than the OWT baseline and ERM-trained SSAEs, indicating their explanations are more effective in predicting activation on new examples. 
Interestingly, despite superior downstream perplexity vs. $L_0$, ERM-trained SSAEs did not yield more interpretable explanations than the baseline.
This aligns with findings in \citet{o2024disentangling}, where interpretability decreased with increasing SAE width attributed to less interpretable fine-grained features. 
As TERM encourages coarser, more compositional features, its explanations are more readily interpretable.

\section{Conclusion and Future Work}
\looseness=-1
This work introduces SSAEs for interpreting rare, subdomain features in FMs. SSAEs trained with Dense retrieval and TERM, outperform standard SAEs in capturing tail concepts and yield more interpretable features. Future work could explore their application to targeted concept unlearning.

\section{Acknowledgements}

This research was supported by the Anthropic Researcher Access Program through their generous grant of model credits, and AI Safety Support. The project originated during Aashiq's participation in the ML Alignment and Theory Scholars (MATS) program. We are grateful to Jake Mendel, Lucius Bushnaq, and Jacob Drori for their insightful discussions on experimental design and valuable feedback on earlier drafts of this manuscript.

\section*{Limitations}

While our work demonstrates the effectiveness of SSAEs in enhancing interpretability and tail concept capture across diverse domains like Physics and Toxicity, there are several areas for further exploration:

\paragraph{Computational Efficiency of TERM.}
\looseness=-1
Training SAEs with TERM, while effective in enhancing concept recall and yielding more interpretable features, can be computationally more demanding than standard ERM. The TERM objective requires computing the exponent of the loss for each data point, which is more computationally intensive than in ERM. This can potentially lead to numerical instability and slower convergence, particularly at high tilt values. 
The benefits of TERM in improving interpretability and fairness encourage further research to reduce its computational cost for broader adoption and scalability.

\paragraph{Dependence on Seed Data.}
\looseness=-1
The success of SSAEs relies on the quality and representativeness of the initial seed data used for retrieval. Low-quality or unrepresentative seed data could lead to SSAEs that fail to capture relevant subdomain concepts or exhibit biases inherited from the seed data. Exploring methods for automatically selecting or generating high-quality seed data and analyzing the sensitivity of SSAEs to different seed data selection strategies would be valuable directions for future research. 

\paragraph{Generalizability Across Domains and Applications.}
\looseness=-1
Our experiments with the Physics, Toxicity, Bias in Bios, and TinyStories datasets demonstrate the effectiveness of SSAEs across diverse domains. While we have no reason to believe our findings won't generalize, further empirical validation across an even broader range of tasks and datasets would strengthen our conclusions. We are particularly interested in evaluating SSAEs in settings where rare concepts play a crucial role, such as AI safety, healthcare, and fairness. These applications would further solidify SSAEs as powerful and versatile tools for enhancing interpretability and control in foundation models.

\section*{Ethical Considerations}

\looseness=-1
The ability to interpret and analyze rare concepts within foundation models, particularly those related to sensitive attributes, carries significant ethical implications that warrant careful consideration.

\paragraph{Potential for Misuse and Dual-Use Concerns.}
\looseness=-1
The techniques presented in this work, while intended for enhancing interpretability, safety, and fairness, could be misused for malicious purposes. The capability to identify and manipulate rare features, especially those associated with sensitive attributes like gender, race, or political affiliation, could be exploited to amplify existing biases, generate harmful or misleading content, or manipulate model behavior in ways that perpetuate or exacerbate societal inequalities. Addressing these dual-use concerns requires proactive efforts to develop safeguards, promote responsible use guidelines, and engage in open discussions about the potential risks associated with these powerful tools.

\paragraph{Bias Amplification.}
\looseness=-1
While SSAEs aim to improve the representation of rare and potentially underrepresented concepts, they are not inherently immune to bias. Biases present in the underlying foundation model and its training data can be inherited and potentially amplified by SSAEs, even when tailored to focus on specific subdomains or sensitive attributes. Mitigating this risk requires careful attention to data curation, development of robust bias detection and mitigation techniques during both FM and SSAE training, and ongoing monitoring and evaluation of SSAE features to ensure they do not perpetuate or exacerbate existing biases.

\paragraph{Data Privacy and Responsible Use.}
\looseness=-1
The datasets used in this work (OWT, Pile, arXiv Physics, Pile Toxicity, Bias in Bios, TinyStories) are publicly available and widely used within the NLP research community (see \autoref{sec:datasetdetails}). These datasets have undergone accepted privacy practices at their creation time. We have strictly adhered to the license terms of these datasets, ensuring responsible and ethical handling. We also acknowledge the contributions of the creators and maintainers of the artifacts used in this work 
(\texttt{Gemma-2b}, \texttt{Pythia-70M}, \texttt{SAELens}, and the \texttt{dictionary\_learning} library). We have utilized these artifacts in accordance with their intended use and licensing agreements.

\paragraph{Reproducibility} To ensure reproducibility and facilitate further research, all our code, experiments, and ablations are implemented within the SAELens framework and will be publicly released upon acceptance of this paper.

\bibliography{custom}

\clearpage
\appendix

\section{Related Work}

\label{sec:related-work}

This work intersects with several research areas, including mechanistic interpretability, sparse coding, feature disentanglement, and evaluation methods for Sparse Autoencoders. We contextualize our contributions within this broader landscape.

\subsection{Mechanistic Interpretability}

\looseness=-1
Mechanistic Interpretability (MI) aims to decipher the internal workings of neural networks by reverse engineering their computational processes \citep{olah2020zoom, elhage2021mathematical}. This approach conceptualizes model computations as collections of circuits – narrow, task-specific algorithms. Recent circuit analyses of Foundation Models (FMs) have focused on mapping these circuits to specific model components like attention heads and MLP layers \citep{wang2022interpretability, heimersheim2023circuit}.

Building upon this component-level understanding, the linear representation hypothesis proposes that component activations can be further decomposed into (sparse) linear combinations of meaningful feature vectors. This concept underpins our work on SSAEs. Unlike previous research that sought to identify individual subspaces representing specific concepts \citep{geiger2023causal, nanda2023emergent, tigges2023linear}, SAEs aim to provide a more complete picture by fully decomposing activations into interpretable features.

MI has shown promise in various downstream tasks, including modifying model behavior to remove toxic outputs \citep{li2023circuit}, altering encoded factual knowledge \citep{meng2022locating}, improving truthfulness \citep{li2024inference}, analyzing gender bias mechanisms \citep{vig2020causal}, and mitigating spurious correlations \citep{gandelsman2023interpreting}. Our work with SSAEs seeks to advance these applications by providing refined tools for detecting, interpreting, and modifying model behavior, particularly concerning rare or underrepresented concepts.

\subsection{Sparse Coding, Dictionary Learning, and Sparse Autoencoders}

Our work draws inspiration from the foundational concepts of sparse coding with over-complete dictionaries \citep{mallat1993matching} and unsupervised dictionary learning from data \citep{olshausen1996emergence}. These ideas, impactful in image processing \citep{mairal2014sparse}, evolved into the development of sparse autoencoders (SAEs) through their integration with autoencoder architectures \citep{hinton2006reducing, lee2007sparse, le2013building, konda2014zero}.

Recently, SAEs have been applied to language models \citep{yun2021transformer, sharkey2023taking, bricken2023monosemanticity, cunningham2023sparse}, with successful implementations on smaller open-source language models \citep{marks2024sparse, bloom2024saetrainingcodebase, mossing2024transformer}. We build upon this research trajectory, addressing specific limitations and extending the approach to capture rare, domain-specific features more effectively.

\subsection{Challenges, Improvements, and Evaluation of Sparse Autoencoders}

Despite their potential, SAEs face several challenges. For example, \citet{anders2024contextlength} observed that SAE features trained on language models with specific context lengths fail to generalize to activations from longer contexts. \citet{wright2024addressing} and \citet{jermyn2024tanh} osberved feature suppression, a phenomenon where SAE feature activations systematically underestimate true activation values due to sparsity penalties.

Various solutions have been proposed to tackle these challenges, including post-training finetuning \citep{wright2024addressing}, alternative sparsity penalties \citep{jermyn2024tanh, riggs2024improving, farrell2024experiments}, and architectural modifications such as Gated SAEs \citep{rajamanoharan2024improving}. Our work focuses on overcoming the limitations of SAEs in representing tail concepts and proposes SSAEs to ensure a more balanced representation of both frequent and rare concepts.

Evaluating SAE performance is further complicated by the absence of ground truth labels for the features they learn. Existing research has employed diverse metrics, including comparison with ground truth features in toy data, activation reconstruction loss, L1 loss, number of alive dictionary elements, feature similarity across seeds and dictionary sizes \citep{sharkey2022taking}, L0 sparsity, KL divergence upon causal interventions \citep{cunningham2023sparse}, reconstructed negative log likelihood \citep{cunningham2023sparse, bricken2023monosemanticity}, feature interpretability \citep{bills2023language}, and task-specific comparisons \citep{makelov2024towards}.

Our work utilizes a combination of these metrics, including L0 sparsity, reconstruction error, downstream perplexity, and automated interpretability evaluations. We also introduce new metrics specifically designed to assess the effectiveness of SSAEs in capturing rare, domain-specific concepts.

\subsection{Disentangled Representations}

Our research also connects to the broader field of disentanglement in representation learning \citep{bengio2013deep}. While traditional disentanglement methods often rely on enforcing priors on learned representations \citep{chen2018isolating, kim2018disentangling, mathieu2019disentangling}, SAEs aim to decompose the representation space of a pretrained language model into a sparse linear combination of an overcomplete basis. This approach aligns with the theory that language models implicitly learn disentangled representations of data with specific structures, which we seek to recover using sparse autoencoders.

\section{Evaluating SSAE for Physics on OOD data}
\begin{figure}[htb]
    \centering
    \includegraphics[width=0.48\textwidth]{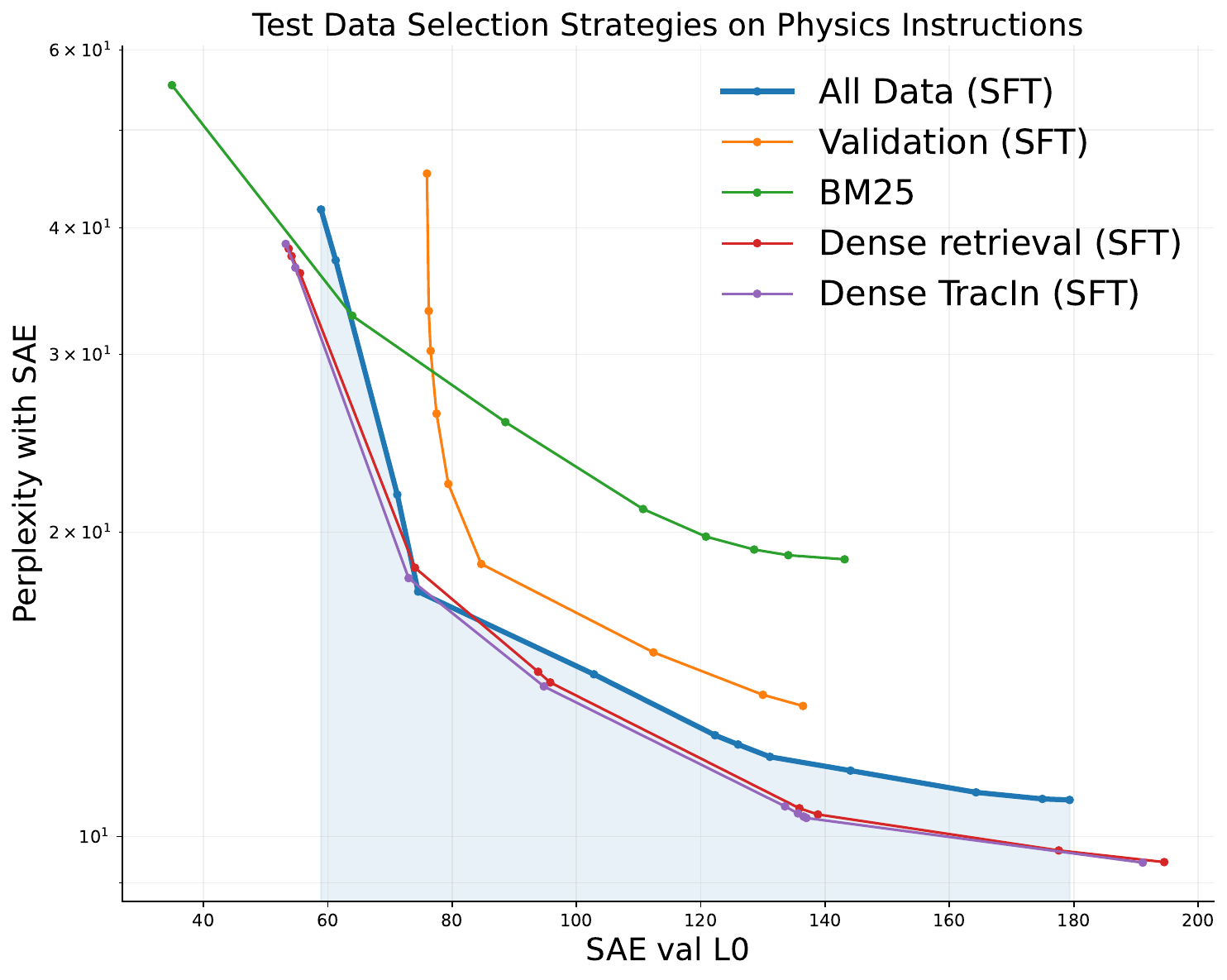}
        \caption{\small Pareto curves for SSAE trained with various data selection strategies as the sparsity coefficient is varied on Physics instruction test data. We plot absolute perplexity with the spliced in SSAE. We find that both BM25 retrieval and training on the validation data generalize poorly when tested out of domain. All curves are averaged over three SAE training run seeds.}
    \label{fig:physicsood}
\end{figure}

\autoref{fig:physicsood} depicts Pareto curves for SSAE trained with various data selection strategies as the sparsity coefficient is varied on the OOD Physics instruction test data. We find that both BM25 retrieval and training on the validation data generalize poorly when tested out of domain.

\section{Evaluating Data Selection Strategies for Toxicity SSAEs }
\label{sec:dataselectiontoxicity}

\begin{figure}[tb]
    \centering
    \subfigure[]{
        \includegraphics[width=\columnwidth]{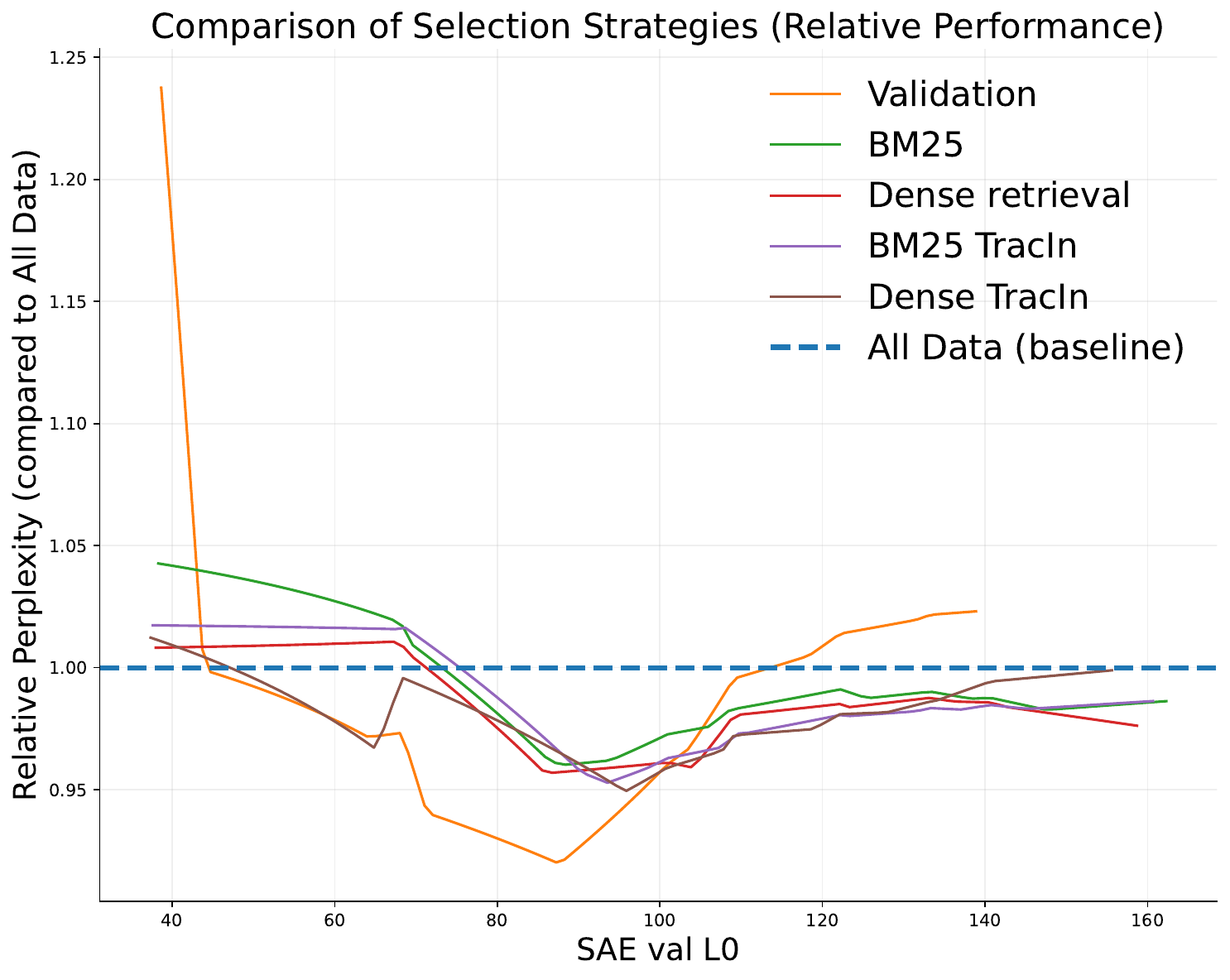}
        \label{fig:toxicityA}
    }
    
    \subfigure[]{
        \includegraphics[width=\columnwidth]{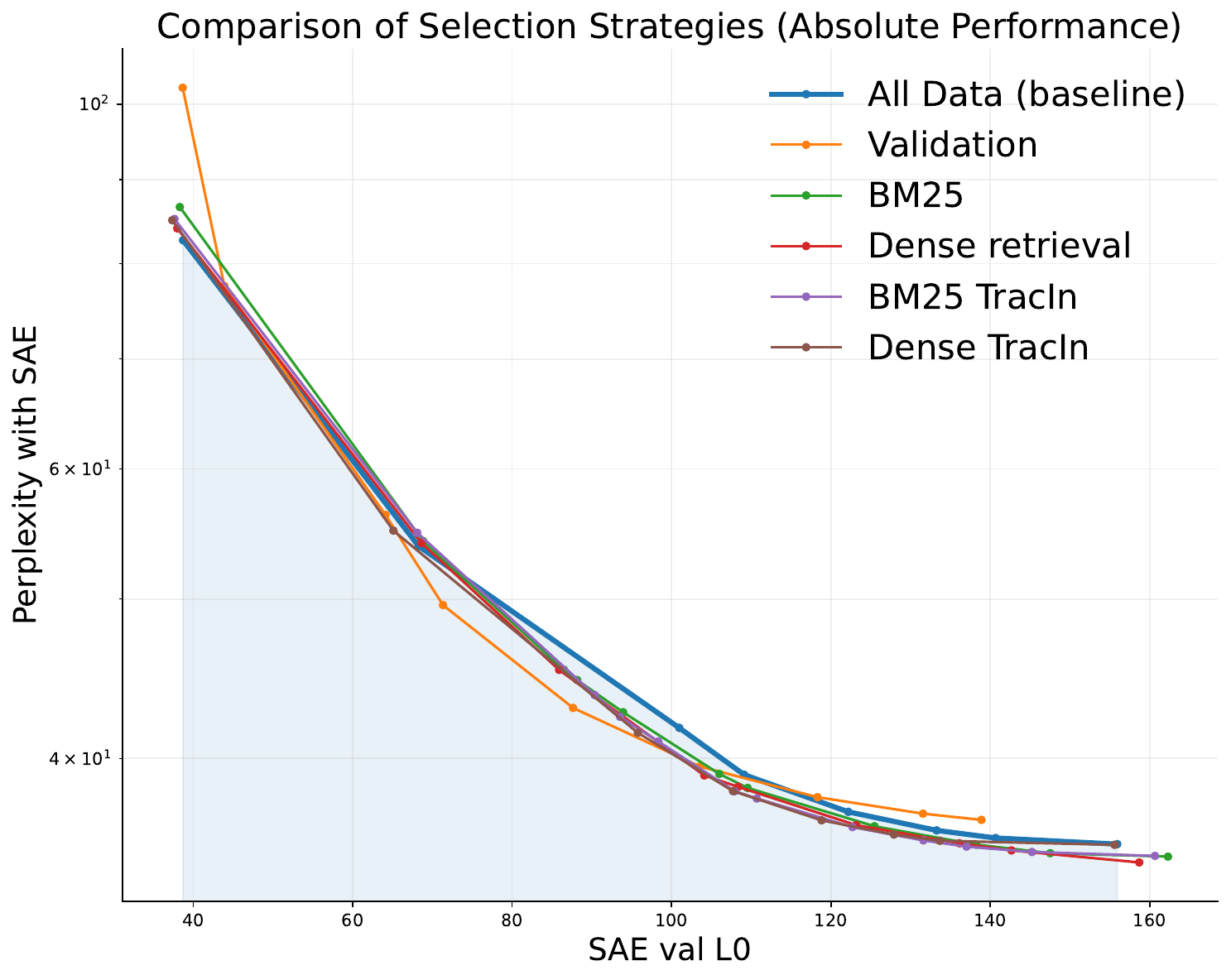}
        \label{fig:toxicityB}
    }
    \caption{\small Pareto curves for Toxicity SSAE trained with various data selection strategies as the sparsity coefficient is varied on Pile toxicity test data. We plot (a) Perplexity with spliced in SAE relative to a GSAE (Baseline) (b) Absolute Perplexity with the spliced in SSAE. Dense TracIn achieves the best performance, followed by Dense retrieval, BM25 TracIn, BM25 and OWT baseline. All curves are averaged over three SAE training run seeds.}
    \label{fig:toxicity}
\end{figure}

We use a seed concept dataset of 4072 tokens from the Pile Toxicity dataset \citep{huggingfacePile}. We retrieve 5.25M tokens from OWT using the same strategies as before and train SSAEs on this data for 500 iterations. We then evaluate the models on a test split of 3.357M tokens from the Pile Toxicity dataset (in-distribution). 
Appendix Figure \ref{fig:toxicity} displays the patched perplexity versus $L_0$ curves for these experiments. The results largely align with the physics experiment, with Dense retrieval outperforming BM25 and TracIn offering a marginal improvement over Dense retrieval alone.

\section{Probing SSAE Tail Concept Learning for Toxicity}

\begin{figure}[htbp]
    \centering
    \includegraphics[width=0.48\textwidth]{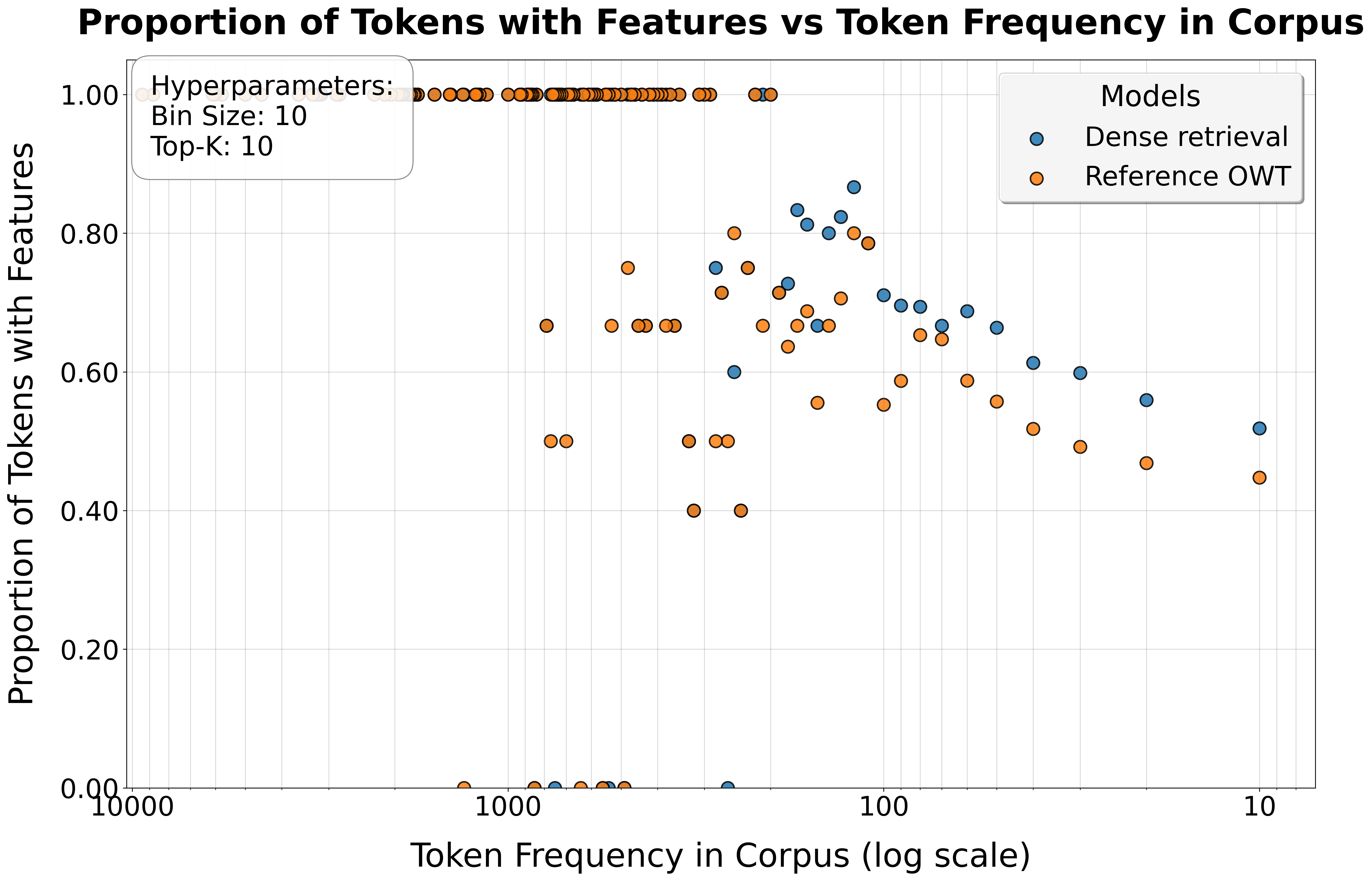}
    \caption{\small Proportion of tokens with SAE features vs. Token frequency in Toxicity data. SSAE trained with dense retrieval captures more tail tokens (concepts) in its features.}
    \label{fig:tailtoxicA}
\end{figure}

\autoref{fig:tailtoxicA} shows the proportion of tokens with SAE features vs. Token frequency in Toxicity data using the Logit Lens approach. We leverage the unembedding matrix as a logit lens to analyze the top-10 token logits associated with each SSAE feature. For each frequency bucket in the Toxicity dataset, we calculate the percentage of tokens that appear among the top-10 logits for at least one feature. This analysis allows us to assess the extent to which SSAE features represent tokens across different frequency ranges. SSAE trained with dense retrieval captures more tail tokens (concepts) in its features compared to the baseline.

\section{Pareto curves for Tilted ERM trained SSAE}
\label{sec:tiltpareto}
\begin{figure}[tb]
    \centering
    \includegraphics[width=0.5\textwidth]{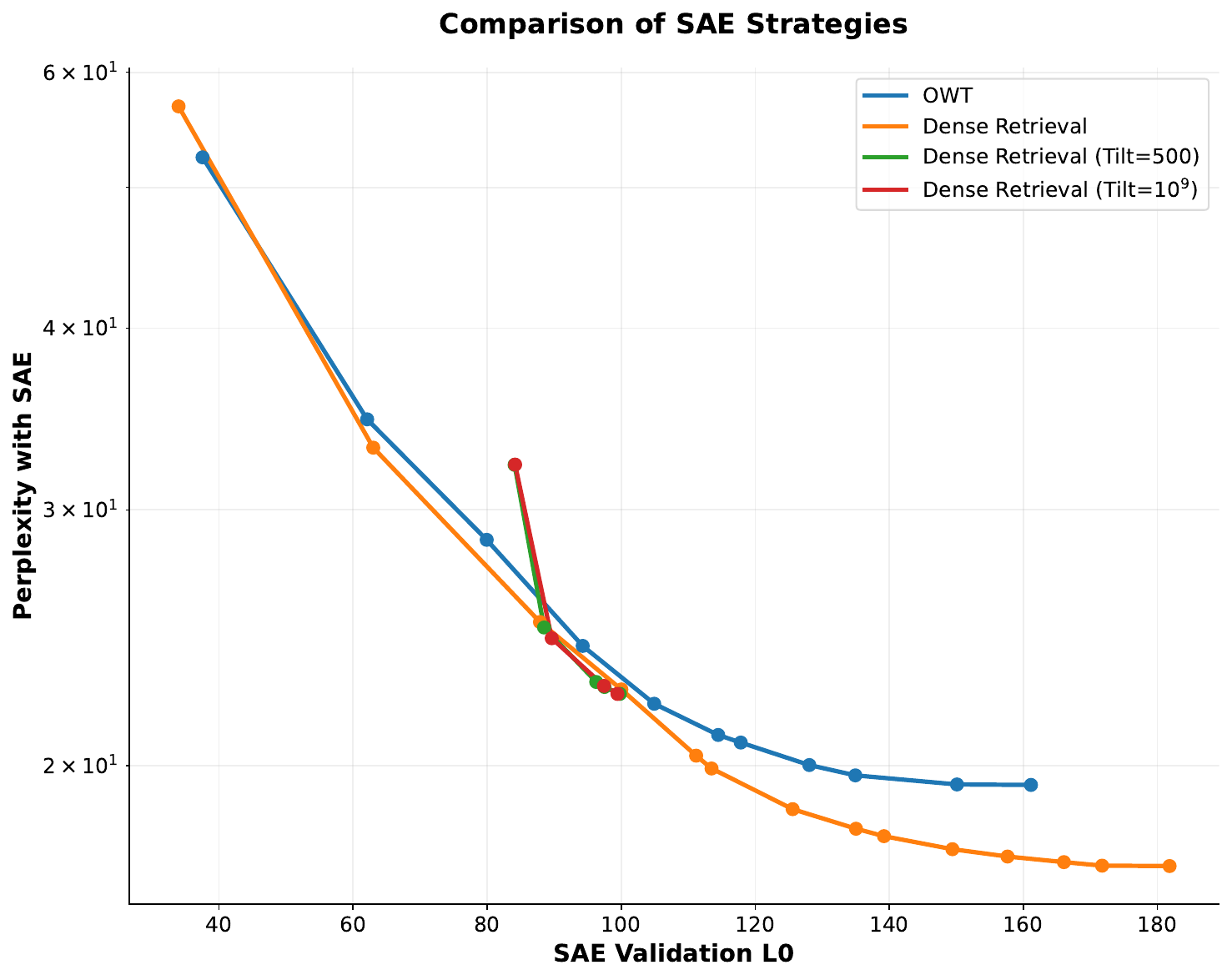}
    \caption{\small Pareto curves for SSAEs finetuned on the \textbf{Physics arXiv dataset} using different strategies: full OpenWebText (OWT), Dense retrieval, and Dense retrieval with Tilted Empirical Risk Minimization (TERM, tilt=500 and TERM, tilt=$10^9$). TERM-finetuned SSAEs achieve competitive performance with Dense retrieval alone within the $L_0$ range of 85-100.  Outside this range, our current training methodology results in higher reconstruction errors. All curves are averaged over three SAE training run seeds.}
    \label{fig:tilteval}
\end{figure}

Figure \ref{fig:tilteval} evaluates SSAEs trained with Tilted ERM on the Physics arXiv dataset, displaying Pareto curves where the x-axis represents $L_0$ and the y-axis shows downstream perplexity with patched-in SSAE. TERM-finetuned SSAEs achieve competitive performance with Dense retrieval alone within the $L_0$ range of 85-100.

\begin{figure}[htb]
    \centering
    \includegraphics[width=0.5\textwidth]{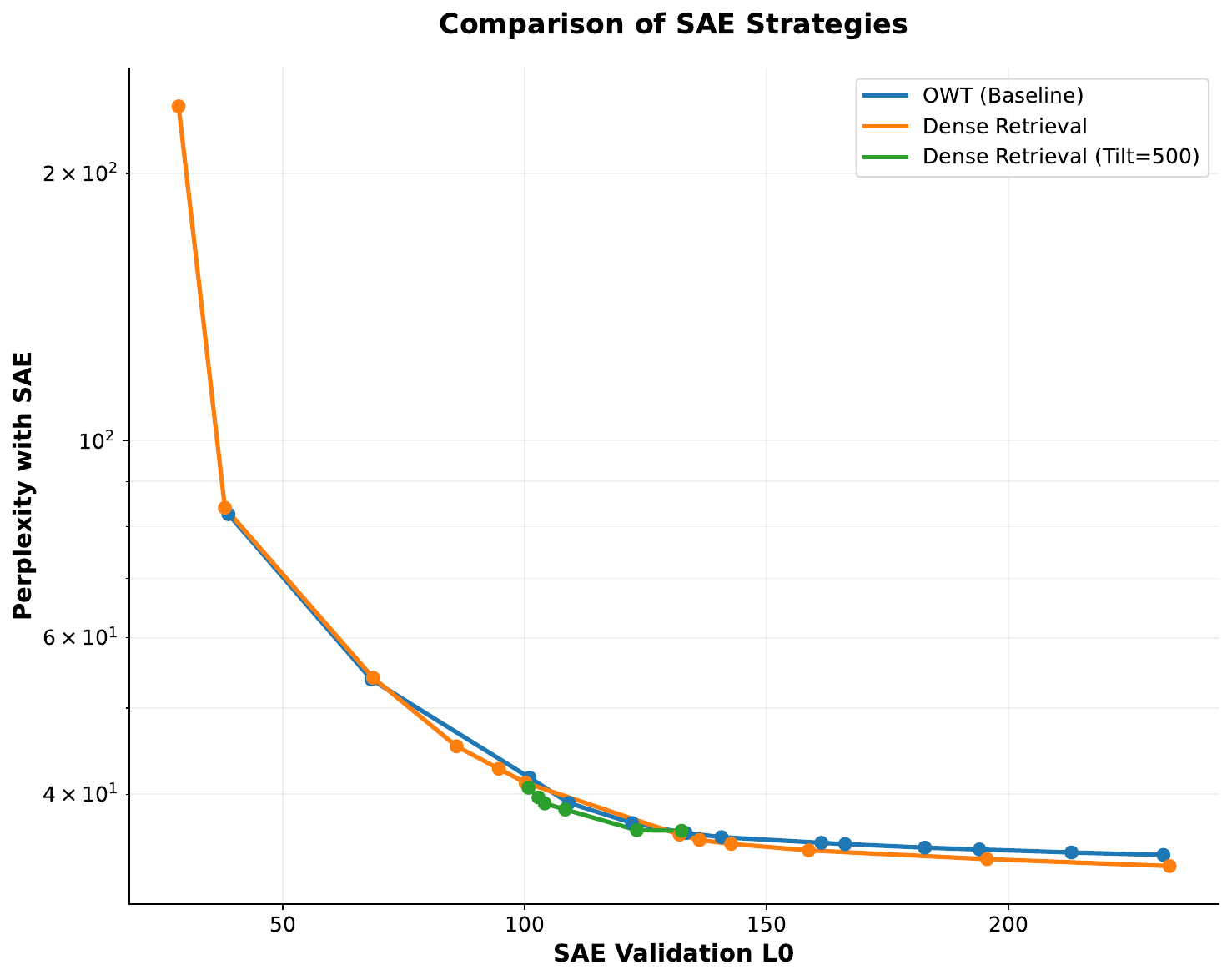}
    \caption{\small Pareto curves for SSAEs finetuned on the \textbf{Toxicity dataset} using different strategies: full OpenWebText (OWT), Dense retrieval, and Dense retrieval with Tilted Empirical Risk Minimization (TERM, tilt=500). TERM-finetuned SSAEs achieve competitive performance with Dense retrieval alone within the $L_0$ range of 100-140. All curves are averaged over three SAE training run seeds. }
    
    \label{fig:tiltoxicteval}
\end{figure}

\autoref{fig:tiltoxicteval} shows similar Pareto curves on the Pile toxicity dataset where TERM-finetuned SSAEs achieve competitive performance with Dense retrieval within the $L_0$ range of 100-140.

Our experiments demonstrate that TERM-trained SSAEs consistently maintain $L_0$ within this desired range, ensuring both sparsity and accurate reconstruction of subdomain concepts.

\paragraph{Improving $L_0$ Control at Extreme Values}
Adaptive penalty schemes are much better than Adam at precisely controlling $L_0$ at extreme values. This approach dynamically adjusts the sparsity penalty $\lambda$ during training based on the current $L_0$. We found that increasing $\lambda$ when $L_0$ exceeds a target range and decreasing it when $L_0$ falls below helped maintain the desired level of sparsity across a wider range of $L_0$ values. This also prevented the emergence of inactive features at low $L_0$ values.

\section{TERM-trained SSAE enhances Tail Concept Capture in Toxicity data}

\begin{figure}[htb]
    \centering
    \includegraphics[width=0.5\textwidth]{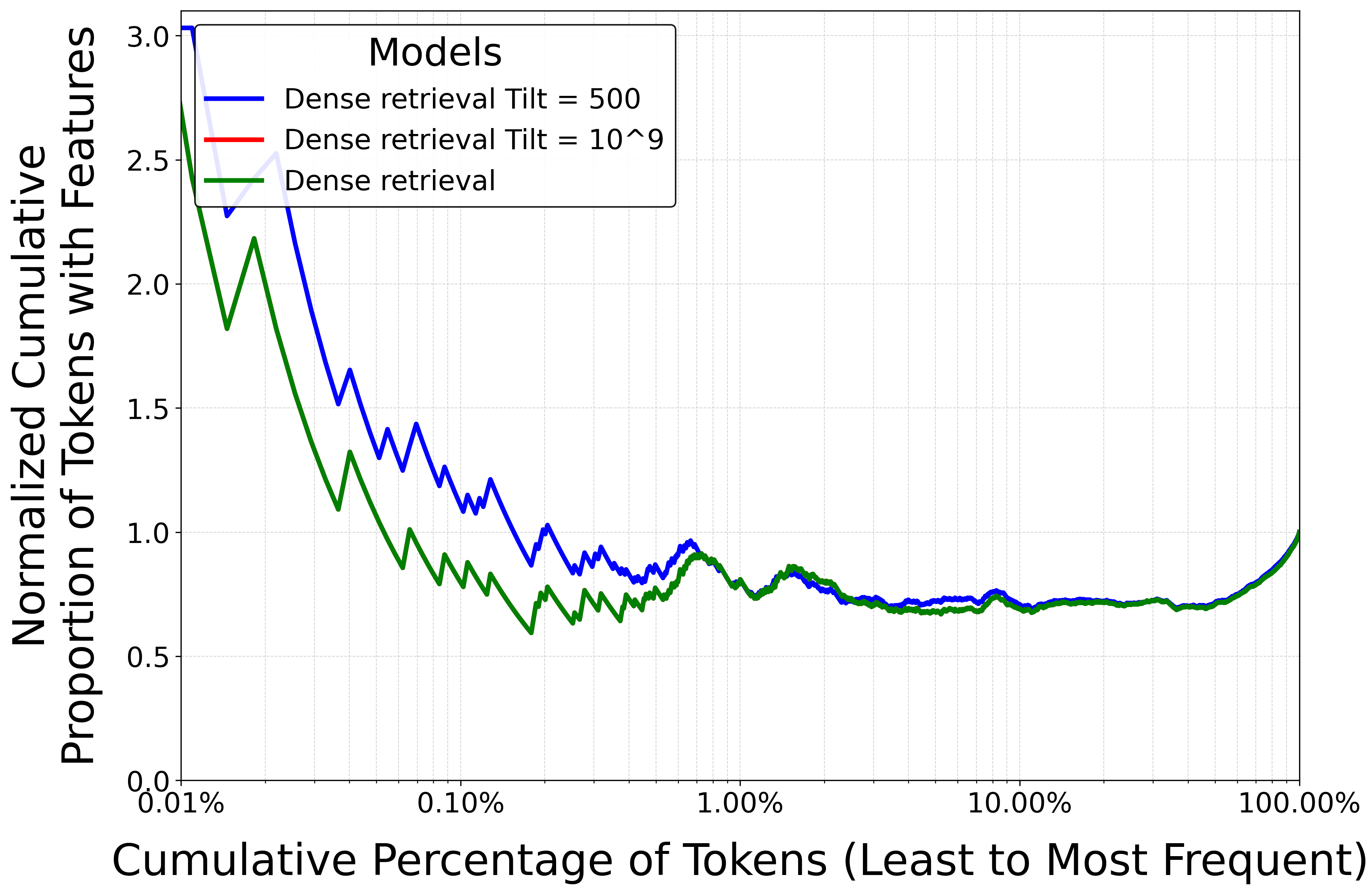}
    \looseness=-1
    \caption{\small Cumulative proportion of tokens with SAE features vs. cumulative percentage of tokens in Toxicity data, normalized per model so that the cumulative proportion of tokens with features is 1 over the entire dataset. SSAE trained with dense retrieval and larger tilt captures more tail tokens (concepts) in its features. Note that the curves at tilt 500 and tilt $10^9$ overlap.}
    \label{fig:tailtoxicB}
\end{figure}

\autoref{fig:tailtoxicB} shows the cumulative proportion of tokens with SAE features vs. cumulative percentage of tokens in Toxicity data, normalized per model so that the cumulative proportion of tokens with features is 1 over the entire dataset. SSAE trained with dense retrieval and larger tilt captures more tail tokens (concepts) in its features.

\section{Implementation Details for Bias-in-Bios Classification Experiments}

We follow the methodology in \citet{marks2024sparse} for  Spurious Human-interpretable Feature Trimming (SHIFT), which we summarize here for completeness. All models can be trained on a single A100 in under a day.

\subsection{Classifier Training}
\label{sec:detailsshift}
Here we describe our approach to training a classifier on Pythia-70M for the Bias in Bios (BiB) task. To mimic a realistic application setting, we conducted a hyperparameter search to train high-performing baseline and oracle classifiers (using the ambiguous and balanced datasets, respectively). Hyperparameters were not selected with the aim of strong SHIFT performance.

\looseness=-1
The inputs to our classifier are residual stream activations from the penultimate layer of Pythia-70M. We apply mean-pooling over (non-padding) tokens from the context. In our initial experiments, we found that extracting representations over only the final token led to slightly worse baseline and oracle performance. Similarly, using activations from Pythia-70M's final layer yielded slightly poorer results.

We then fit a linear probe to these representations using logistic regression. For optimization, we employ AdamW \citep{loshchilov2017decoupled} with a learning rate of 0.01, training for a single epoch. When retraining after SHIFT, we finetune only this linear probe, leaving the full model unchanged.

Like \citet{marks2024sparse}, we encountered difficulties when attempting to fit a probe with greater-than-chance accuracy using logistic regression on final layer representations. This observation led us to opt for penultimate layer representations in our main approach.

\subsection{Implementation for Concept Bottleneck Probing}
\label{sec:detailscbp}
Our implementation of Concept Bottleneck Probing (CBP) draws from \citet{yan2023robust}. The process is as follows:

\begin{enumerate}[itemsep=0pt, topsep=0pt, leftmargin=10pt, rightmargin=1pt, labelsep=2pt, itemindent=2pt, parsep=0pt]
    \item First, we select $N = 20$ keywords related to the intended prediction task. Our keyword set includes: nurse, healthcare, hospital, patient, medical, clinic, triage, medication, emergency, surgery, professor, academia, research, university, tenure, faculty, dissertation, sabbatical, publication, and grant.
    
    \item We obtain concept vectors $c_1, \ldots, c_N$ for each keyword by extracting Pythia-70M's penultimate layer representation over the final token of each keyword, then subtracting the mean concept vector. This normalization step proved crucial, as we found that without it, concept vectors exhibited very high pairwise cosine similarities.
    
    \item Given an input with representation $x$ (obtained via the mean-pooling procedure described earlier), we construct a concept bottleneck representation $z \in \mathbb{R}^N$ by computing the cosine similarity with each $c_i$.
    
    \item Finally, we train a linear probe on these concept bottleneck representations $z$ using logistic regression, following the approach outlined in the Classifier Training subsection.
\end{enumerate}

As in \citet{marks2024sparse}, we decided to normalize concept vectors but not input representations, as this approach yielded stronger performance. We also explored the alternative of computing cosine similarities before mean pooling.

\section{Sparse Feature Circuits for Bias in Bios Classifer}
\label{sec:detailscircuits}

\begin{figure*}[htb]
    \centering
    \includegraphics[width=\textwidth]{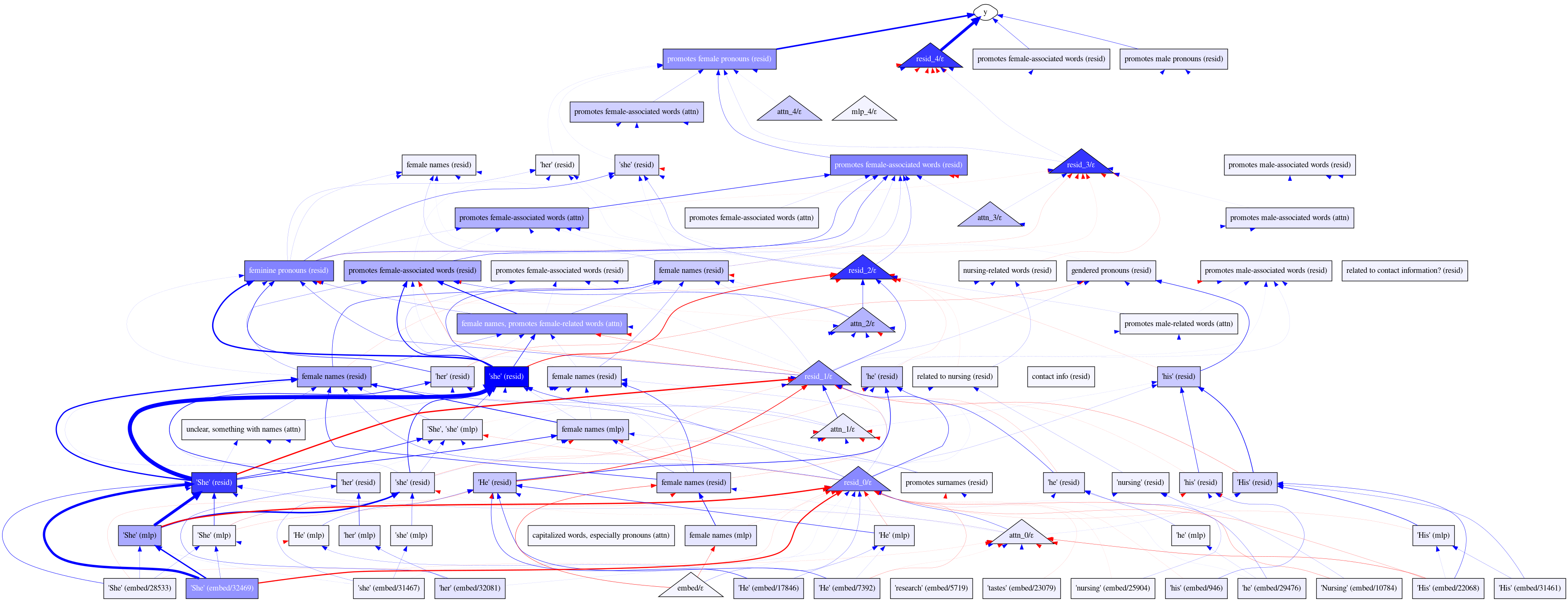}
    \caption{\small The full annotated feature circuit discovered for the Bias in Bios classifier with the \textbf{GSAE patched in}. The circuit was discovered using $T_N = 0.1$ and $T_E = 0.01$. We observe that the circuit contains many nodes that simply detect the presence of gendered pronouns or gendered names. A few features attend to profession information, including one which activates on words related to nursing, and another that activates on passages relating to science and academia.}
    \label{fig:circuit_GSAE}

    \includegraphics[width=\textwidth]{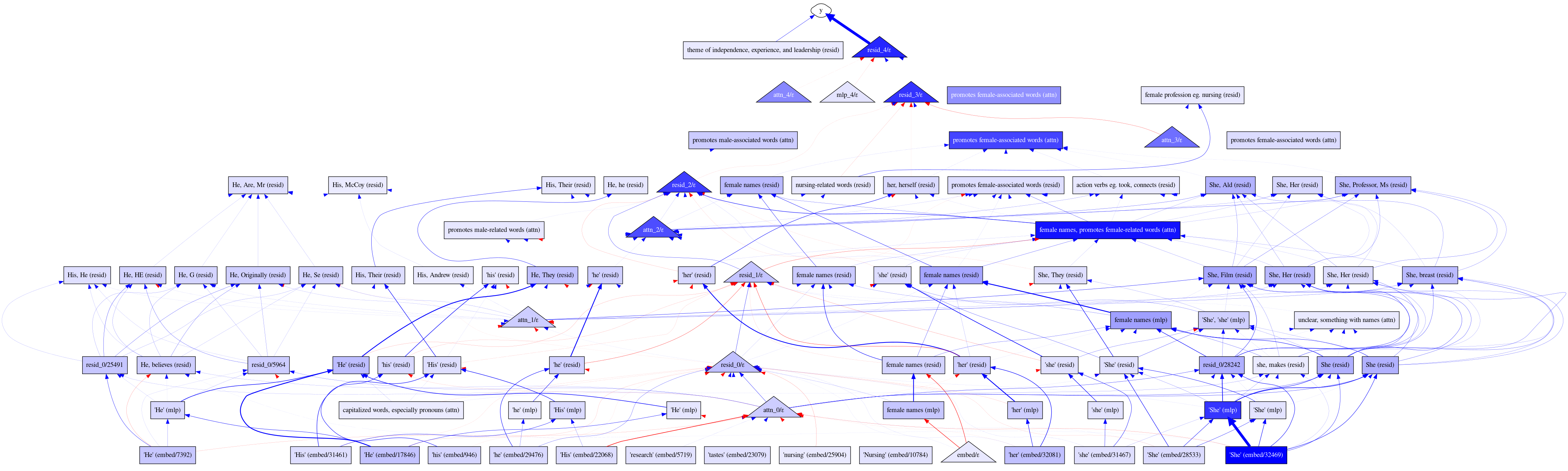}
    \caption{\small \looseness=-1 The full annotated feature circuit for the Bias in Bios classifier with the \textbf{finetuned SSAE patched in}. The circuit was discovered using $T_N = 0.1$ and $T_E = 0.01$. This circuit is much larger due to newly activated features in the SSAE that detect the presence of gendered pronouns and gendered names, as well as features for profession information such as nursing and academia.}
    \label{fig:circuit_SSAE}
\end{figure*}

In this section, we generate sparse feature circuits, which are computational sub-graphs that explain model behaviors in terms of SAE features and error terms, using the methodology in \citet{marks2024sparse}. We begin by describing the process of generating these circuits.

Given a language model $M$, SAEs for various submodules of $M$ (e.g., attention outputs, MLP outputs, and residual stream vectors), a dataset $D$ consisting of either contrastive pairs $(x_\text{clean}, x_\text{patch})$ of inputs or single inputs $x$, and a metric $m$ depending on $M$'s output when processing data from $D$, we can construct these circuits. The idea is to treat SAE features as part of the model. By applying the decomposition to various hidden states $x$ in the LM, we can view the feature activations $f_i$ and SAE errors $\varepsilon$ as integral parts of the LM's computation. This allows us to represent the model as a computation graph $G$ where nodes correspond to feature activations or SAE errors at particular token positions.

To approximate the Indirect Effect (IE) of each node, we compute $\hat{IE}(m; a; x)$ for each node $a$ in $G$ and input $x \sim D$, where $\hat{IE}$ is either $\hat{IE}_\text{atp}$ or $\hat{IE}_\text{ig}$. We then apply a node threshold $T_N$ to select nodes with a large (absolute) IE. Consistent with prior work \citep{nanda2023attribution, kramar2024atp}, we find that $\hat{IE}_\text{atp}$ accurately estimates IEs for SAE features and errors, except for nodes in the layer 0 MLP and early residual stream layers. For these components, $\hat{IE}_\text{ig}$ significantly improves accuracy, so we employ it in our experiments.

We also compute the average IE of edges in the computation graph using an analogous linear approximation. After computing these IEs, we filter for edges with absolute IE exceeding some edge threshold $T_E$.

For templatic data where tokens in matching positions play consistent roles, we take the mean effect of nodes/edges across examples. For non-templatic data, we first sum the effects of corresponding nodes/edges across token positions before taking the example-wise mean \citep{marks2024sparse}.

Figure \ref{fig:circuit_GSAE} presents the full annotated feature circuit for the Bias in Bios linear classifier based on Pythia-70M with the pretrained GSAE patched in. The annotations are from human inspection of examples that activate features. Many nodes simply detect the presence of gendered pronouns or gendered names. A few features attend to profession information, including one which activates on words related to nursing, and another which activates on passages relating to science and academia.

\looseness=-1
Similarly, Figure \ref{fig:circuit_SSAE} displays the full annotated feature circuit for the Bias in Bios linear classifier based on Pythia-70M with the finetuned SSAE patched in. This circuit, discovered using $T_N = 0.1$ and $T_E = 0.01$, is much larger due to newly activated features in the SSAE that detect the presence of gendered pronouns and gendered names, as well as features for profession information such as nursing and academia. This is responsible for the improved classification performance with the SSAE.

In each circuit, sparse features are shown in rectangles, whereas causally relevant error terms not yet captured by our SAEs are shown in triangles. Nodes shaded in darker colors have stronger effects on the target metric $m$. Blue nodes and edges are those which have positive indirect effects (i.e., are useful for performing the task correctly), whereas red nodes and edges are those which have counterproductive effects on $m$ (i.e., cause the model to consistently predict incorrect answers).

\section{Relative Feature Activation Distribution}

\begin{figure}[htb]
    \centering
    \includegraphics[width=0.48\textwidth]{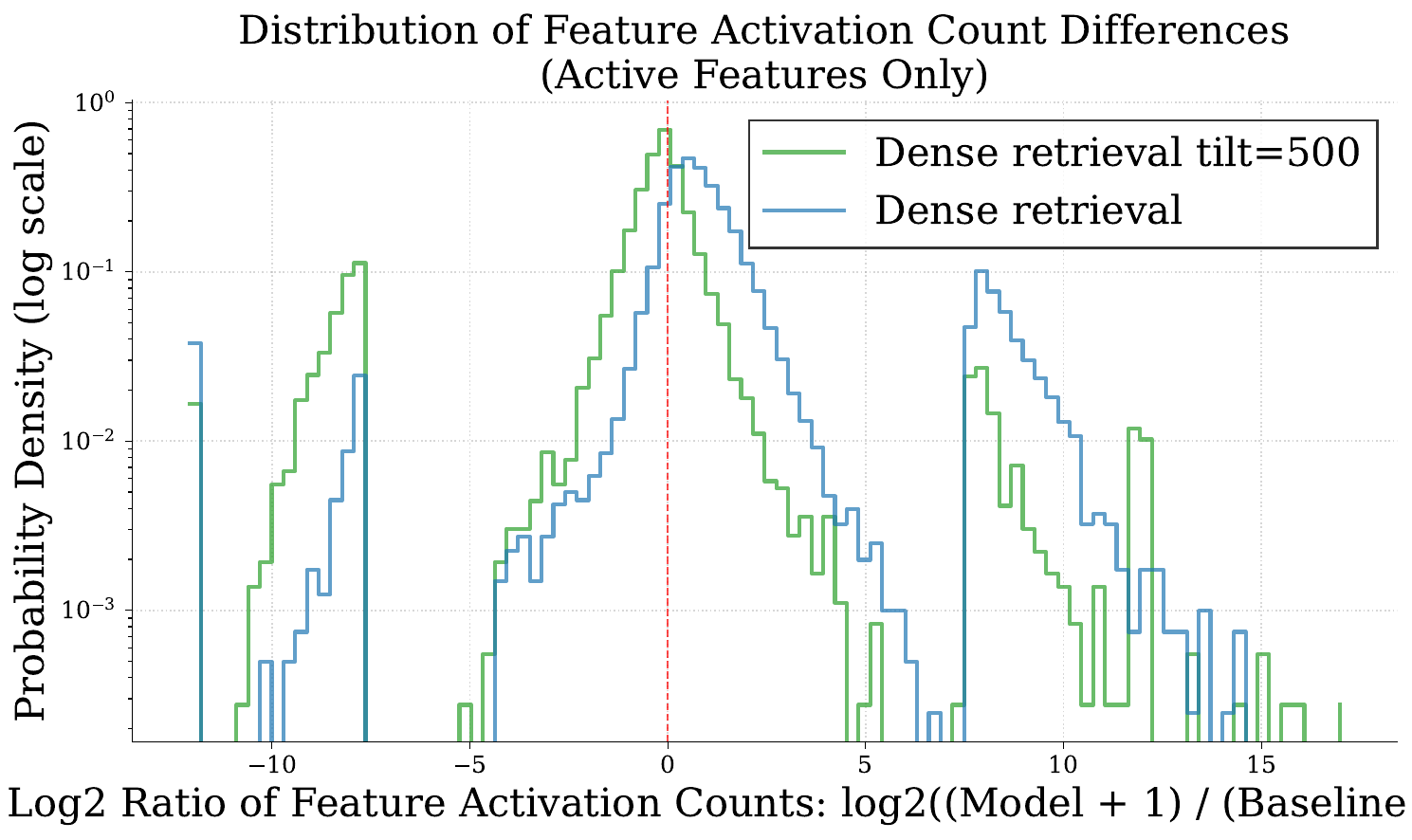}

    \caption{\small Distribution of log-ratio feature activation count differences between specialized SAEs and the OWT baseline on the Physics arXiv test set, normalized per SAE model. Blue represents the ERM-trained SSAE with Dense retrieval, orange represents the TERM-trained SSAE with tilt=500. The ERM-trained SSAE exhibits more probability mass on the right, indicating an emphasis on representing common concepts, while the TERM-trained SSAE's shift towards the left suggests a greater focus on representing domain-specific tail concepts.}
    \label{fig:relativehistogram2}
\end{figure}

\begin{figure}[htb]
    \centering
    \includegraphics[width=0.48\textwidth]{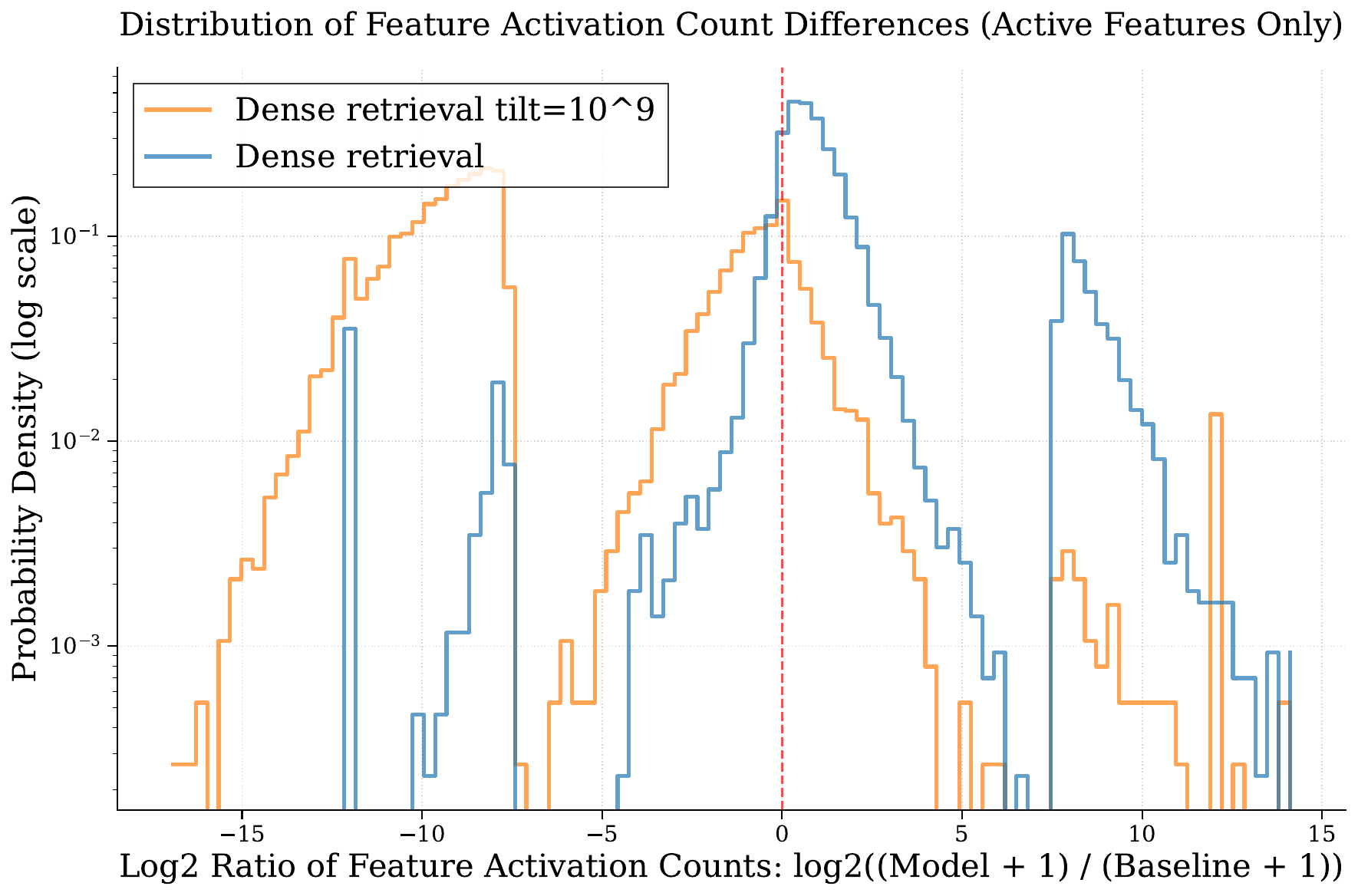}

    \caption{\small Distribution of log-ratio feature activation count differences on the Physics arXiv test set, normalized per SAE model. Blue represents the ERM-trained SSAE with Dense retrieval, orange represents the TERM-trained SSAE with tilt=$10^9$. The intensified leftward shift of probability mass with higher tilt demonstrates that TERM increasingly prioritizes representing tail concepts compared to standard ERM-trained SSAE, which focuses more on frequent concepts.}
    \label{fig:relativehistogram3}
\end{figure}

Figures \ref{fig:relativehistogram2} and \ref{fig:relativehistogram3} analyze the distribution of differences in feature activation counts between the same features in specialized SAEs (both ERM and TERM-trained) and the OWT baseline on the Physics arXiv test set. The difference is quantified as the log ratio of feature activation counts: $\log_2 (\frac{M+1}{B+1})$, where $M$ represents the SSAE and $B$ the OWT baseline. Positive values indicate features activating on more data points in the specialized SAEs relative to the baseline SAE.

Finetuning on the subdomain with ERM  leads to an increase in feature activation counts overall, as evidenced by the positive probability mass. This adaptation reflects the SSAE features specializing towards concepts prevalent in the Physics arXiv dataset.

Training SSAEs with TERM, which minimizes worst-case performance, distinctly alters feature activation patterns. Compared to standard ERM, TERM-trained SAEs concentrate more probability mass on the distribution's left side, indicating many features are less activated relative to the baseline. This leftward shift aligns with the theoretical underpinnings of TERM, which encourages robustness to distribution shift and tail events. By upweighting worse-performing examples, TERM promotes the activation of features crucial for capturing tail concepts. The TERM-trained SAE redistributes its capacity, with numerous features specializing in tail concepts (low-level activations), while others become more general activating on a wider range of concepts. This shift towards negative relative counts intensifies with increasing tilt, suggesting that higher tilt values further prioritize the representation of tail concepts.

\section{Automated Intepretability Explanations}
\label{sec:explanations}

\looseness=-1
Boxes~\ref{tcb:explanation_gsae}, \ref{tcb:explanation_ssae}, and \ref{tcb:explanation_tilted_ssae} show the Interpreter's explanations for the active features among the first ten features (by count) of the pretrained GSAE, the ERM-trained SSAE, and the TERM-trained SSAE, respectively, on the arXiv Physics test set. We observe a clear distinction in how these models specialize and represent concepts. While the ERM-trained SSAE activates more features than the GSAE, reflecting its focus on frequent concepts within the domain, its explanations are more complex and less readily interpretable. Conversely, the TERM-trained SSAE, despite activating fewer features overall, produces explanations that are easier to understand. This suggests that TERM learns features that are compositional and encourages a balanced representation of both frequent and rare concepts. The lower number of active features for the TERM-trained SSAE could be attributed to the potential absence of many tail concepts in the test set.

\begin{figure*}[htbp]
\begin{tcolorbox}[
  colback=blue!5!white,
  colframe=blue!75!black,
  title= Box 1: Generalized SAE,
  fontupper=\small,
  width=\textwidth,
  parbox=false
]
\refstepcounter{mytcb}\label{tcb:explanation_gsae}

0. The token "0" appearing in scientific notation, journal article citations, or encoded ASCII representations, often in the context of physics or chemistry literature references.

5. This neuron appears to activate on mathematical and scientific notation, particularly symbols, equations, and specialized formatting in technical documents. It may play a role in recognizing and processing scientific or mathematical content within text.

7. The neuron appears to activate on punctuation marks, particularly commas and quotation marks, when they are used to separate or enclose items in a list, mathematical expressions, or technical notation in scientific or mathematical text. It may play a role in parsing and understanding the structure of complex technical writing.

8. This neuron appears to activate on tokens that are part of or follow noun phrases, often in technical or academic contexts. It seems to be sensitive to words that introduce or refer to specific objects, concepts, or pieces of information within a larger text. The neuron may play a role in tracking referential elements or key pieces of information in complex, information-dense text.

9. The token "," appearing after complex scientific or technical phrases, often preceding conjunctions or additional clauses that provide further explanation or context in academic or scientific writing.

10. This neuron appears to activate on abbreviated references to academic or scientific sources, particularly in bibliographies or citation lists. It responds to: 1. Abbreviated journal names (e.g. "NY", "APS", "Euro") 2. Abbreviated organization names (e.g. "SIAM", "INSPEC") 3. URL components of online references (e.g. "citeseer", "philsci", "biology-") 4. Abbreviated publisher names (e.g. "TERRAPUB") The neuron seems to play a role in recognizing citation patterns.
\end{tcolorbox}

\begin{tcolorbox}[
  colback=green!5!white,
  colframe=green!75!black,
  title=Box 2: Specialized SAE,
  fontupper=\small,
  parbox=false
]
\refstepcounter{mytcb}\label{tcb:explanation_ssae}

0. The token "0" appearing in scientific paper citations, journal volume numbers, or ASCII code representations, often in the context of physics or mathematics literature.

4. This neuron appears to activate on tokens related to academic and scientific writing, particularly in the context of physics, science education, and the philosophy of science. It frequently activates on words like "universities", "science", "class", "theories", and other academic terminology. The neuron may be involved in recognizing and generating text related to scientific discourse and academic writing.

5. This neuron appears to activate on scientific and mathematical notation, particularly superscripts, subscripts, and special characters used in equations and formulas. It may play a role in processing and understanding technical or scientific text.

7. The token "by" often appears before introducing a variable, parameter, or label in mathematical or scientific text. It is frequently used to define or denote specific elements in equations, models, or experimental setups.

8. The neuron appears to activate on numerical digits, particularly the digit "4", within scientific or technical contexts such as citations, measurements, or equipment specifications. This suggests the neuron may play a role in identifying or processing numerical information in academic or technical writing.

9. The token "," after various phrases in scientific or technical writing, often used to separate clauses or elements in a list. This neuron may be detecting punctuation patterns in formal, academic-style text.

10. This neuron appears to activate on abbreviations and short identifiers in academic or scientific references, particularly those related to publications, databases, or online resources. Examples include "cites", "NY", "ZIN", "TER", "SI", "e-", "cond", "Compustat", "ASP", "IN", "CAS", "Physics", "Pren", "ourworld", "compuserve", and "APS". These often appear in bibliographic entries, URLs, or other citation-related contexts in academic writing.
\end{tcolorbox}

\begin{tcolorbox}[
  colback=red!5!white,
  colframe=red!75!black,
  title=Box 3: Specialized SAE with Tilt 500,
  fontupper=\small,
  parbox=false
]
\refstepcounter{mytcb}\label{tcb:explanation_tilted_ssae}

0. The token "0" appearing in scientific notation, particularly in journal citations, volume numbers, and page numbers. This neuron may be involved in recognizing and processing numerical information in academic or scientific contexts.

5. This neuron appears to activate on mathematical and scientific notation, particularly equations, variables, and symbols. It seems to be sensitive to complex mathematical expressions, physical constants, and scientific formulas across various fields including physics, chemistry, and engineering. The neuron may play a role in processing and generating technical scientific content.

7. The neuron appears to activate on punctuation marks, particularly commas and angle brackets, when used to separate or enclose items in mathematical or scientific notation. It may play a role in parsing and understanding the structure of technical or mathematical text.

9. The token "," after phrases or clauses, often used to separate elements in scientific or technical writing. This neuron may be detecting punctuation patterns in formal, academic text.
\end{tcolorbox}
\end{figure*}

\section{Automated Interpretability Prompts}
\label{sec:autointerp_prompts}

In this section, we present the Interpreter and Predictor prompts used with Claude 3.5 Sonnet~(\texttt{claude-3-5-sonnet-20240620}) in our automated interpretability pipeline. We note that all AutoInterp experiments cost less than $\$1000$ to run.

\subsection{Interpreter Prompt}

The Interpreter prompt in Box \ref{tcb:interpreter_prompt} is designed to analyze SAE feature activations and explain what causes a specific feature to activate. It is given a list of text examples where the feature activates, with the activating tokens highlighted.

\begin{figure*}[htbp]
\begin{tcolorbox}[colback=gray!5!white,colframe=gray!75!black,title=Box 4: Interpreter Prompt]
\refstepcounter{mytcb}\label{tcb:interpreter_prompt}
\begin{minipage}{\dimexpr\linewidth-2\fboxsep\relax}
\small
\begin{Verbatim}[breaklines=true, breakanywhere=true]
SYSTEM = """You are a meticulous AI researcher conducting an important investigation into a certain neuron in a language model. Your task is to analyze the neuron and explain what causes the neuron to activate.
{prompt}
Guidelines:
You will be given a list of text examples on which the neuron activates. The specific tokens which cause the neuron to activate will appear between delimiters like <<this>>. If a sequence of consecutive tokens all cause the neuron to activate, the entire sequence of tokens will be contained between delimiters <<just like this>>.
- You must produce a concise final description. Simply describe the text features that activate the neuron, and what its role might be based on the tokens it predicts.
- The last line of your response must be the formatted explanation.
- Think carefully about the patterns in the text examples and the tokens that activate the neuron. Pay attention to detail.
{subject_specific_instructions}"""
\end{Verbatim}
\end{minipage}
\end{tcolorbox}

\begin{tcolorbox}[colback=gray!5!white,colframe=gray!75!black,title=Box 5: Interpreter Example, width=\textwidth]
\refstepcounter{mytcb}\label{tcb:interpreter_example}
\begin{minipage}{\dimexpr\linewidth-2\fboxsep\relax}
\small
\begin{verbatim}
EXAMPLE_1 = """
Example 1:  and he was <<over the moon>> to find
Example 2:  we'll be laughing <<till the cows come home>>! Pro
Example 3:  thought Scotland was boring, but really there's more 
<<than meets the eye>>! I'd
"""
EXAMPLE_1_EXPLANATION = """
[EXPLANATION]: Common idioms in text conveying positive sentiment.
"""
\end{verbatim}
\end{minipage}
\end{tcolorbox}
\end{figure*}

\begin{figure*}[htbp]
\begin{tcolorbox}[colback=gray!5!white,colframe=gray!75!black,title=Box 6: Predictor Prompt, width=\textwidth]
\refstepcounter{mytcb}\label{tcb:predictor_prompt}
\begin{minipage}{\dimexpr\linewidth-2\fboxsep\relax}
\small
\begin{verbatim}
DSCORER_SYSTEM_PROMPT = """You are an intelligent and
meticulous linguistics researcher.
You will be given a certain feature of text, such as
"male pronouns" or "text with negative sentiment".
You will then be given several text examples. Your task
is to determine which examples possess the feature.
For each example in turn, return 1 if the sentence is
correctly labeled or 0 if the tokens are mislabeled. You
must return your response in a valid Python list. Do not
return anything else besides a Python list.
"""
\end{verbatim}
\end{minipage}
\end{tcolorbox}

\begin{tcolorbox}[colback=gray!5!white,colframe=gray!75!black,title=Box 7: Predictor Example, width=\textwidth]

\refstepcounter{mytcb}\label{tcb:predictor_example}
\begin{minipage}{\dimexpr\linewidth-2\fboxsep\relax}
\small
\begin{verbatim}
DSCORER_EXAMPLE_1 = """Feature explanation: "of" before words that start
with a capital letter.
Text examples:
Example 0: climate, Tomblinâ Chief of Staff Charlie Lorensen said.
Example 1: no wonderworking relics, no true Body and Blood of Christ,
no true Baptism
Example 2:Deborah Sathe, Head of Talent Development and Production
at Film London,
Example 3: It has been devised by Director of Public Prosecutions (DPP)
Example 4: and fair investigation not even include the Director of
Athletics? Finally, we believe the
"""
DSCORER_RESPONSE_1 = "[1,1,1,1,1]"
\end{verbatim}
\end{minipage}
\end{tcolorbox}
\end{figure*}

\subsubsection{Example Application of Interpreter Prompt}

Box \ref{tcb:interpreter_example} provides an example of how the Interpreter prompt is applied.

\subsection{Predictor Prompt}

The Predictor prompt in Box \ref{tcb:predictor_prompt} is used to predict given a feature explanation whether the given text examples activate the feature. It returns a binary classification label for each example.

\subsubsection{Example Application of Predictor Prompt}

Box \ref{tcb:predictor_example} provides an example of how the Predictor prompt is applied.

\section{Proof of Lower Description Length under Tilted ERM}
\label{sec:proof}
We prove that training a Sparse Autoencoder (SAE) using Tilted ERM leads to a lower total description length compared to standard ERM under specific conditions, suggesting Tilted ERM produces more interpretable features according to the Minimum Description Length (MDL) principle.

\subsection{Problem Setup and Assumptions}

We consider a dataset $\mathcal{D} = \{ x_i \}_{i=1}^N$, where each $x_i \in \mathbb{R}^d$ is generated from a mixture of two Gaussian distributions: a majority cluster (Cluster A) and a minority cluster (Cluster B). Cluster A has mean $\boldsymbol{\mu}_A = \mathbf{0}$, covariance $\Sigma_A = \sigma^2 \mathbf{I}$, and proportion $q_A = N_A/N$. Cluster B has mean $\boldsymbol{\mu}_B = \boldsymbol{\delta}$ (where $\boldsymbol{\delta} = \delta \mathbf{1}$, $\delta > 0$), covariance $\Sigma_B = \sigma^2 \mathbf{I}$, and proportion $q_B = N_B/N = 1 - q_A$. We assume $q_A \gg q_B$, reflecting a significant class imbalance often encountered in real-world scenarios.

The SAE consists of an encoder $h_i = W x_i$ and a decoder $\hat{x}_i = W^\top h_i$, where $W \in \mathbb{R}^{k \times d}$ is the weight matrix and $h_i \in \mathbb{R}^k$ is the latent representation. Sparsity is enforced through an $L_1$ penalty in the loss function, defined as $L(x_i; W) = \| x_i - \hat{x}_i \|^2 + \lambda \| h_i \|_1$, where $\lambda > 0$ controls the trade-off between reconstruction error and sparsity. Assume the nonlinearity is always activated i.e., the identity function.

We compare two training objectives: standard ERM, which minimizes the average loss $\frac{1}{N} \sum_{i=1}^N L(x_i; W)$, and Tilted ERM, which approximates the minimization of the maximum loss through the objective $\frac{1}{\tau} \log (\sum_{i=1}^N e^{\tau L(x_i; W)})$ for large $\tau > 0$.

We make several simplifying assumptions. First, we assume binary latent codes, where $h_{ij} \in \{0, 1\}$. This assumption, while a simplification of continuous-valued activations, allows for a clearer analysis of feature interpretability through the lens of information theory. Second, we assume that features are activated independently, which, while not always true in practice, provides a tractable framework for our analysis. Lastly, we assume uniform activation probabilities across features within each cluster, which simplifies our calculations while still capturing the essential dynamics of the system.

\subsection{Description Length and Feature Activation Probabilities}

The total description length is given by $DL_{\text{total}} = DL_{\text{model}} + DL_{\text{data}}$. Since $DL_{\text{model}}$ is the same for both ERM and Tilted ERM (assuming identical model capacity), we focus our analysis on $DL_{\text{data}}$, which represents the description length of the latent representations $\{h_i\}$.
For a binary latent vector $h_i$, the description length is given by:
\begin{equation}
\scalebox{0.9}{$
\small
\begin{aligned}
DL(h_i) = \sum_{j=1}^k - \biggl( &h_{ij} \log_2 P(h_{ij} = 1) \\[-2ex]
&+ (1 - h_{ij}) \log_2 P(h_{ij} = 0) \biggr)
\end{aligned}
$}
\end{equation}

Given our assumption of independent features, the expected description length per data point from Cluster $C$ ($C \in \{A, B\}$) is:
\begin{small}
\begin{equation}
\begin{gathered}
DL_C = k \cdot H(p_C)
\end{gathered}
\end{equation}
\end{small}
where $p_C$ is the activation probability for features in Cluster $C$, and $H(p)$ is the binary entropy function: $H(p) = - p \log_2 p - (1 - p) \log_2 (1 - p)$.
The total description length for the data is thus:
\begin{equation}
\scalebox{0.9}{$
\begin{aligned}
DL_{\text{data}} &= N_A DL_A + N_B DL_B \\
&= N_A k H(p_A) + N_B k H(p_B)
\end{aligned}
$}
\end{equation}
Our goal is to show that under certain conditions, $DL_{\text{data}}^{\text{Tilted}} < DL_{\text{data}}^{\text{ERM}}$.

\subsection{Analysis of ERM vs. Tilted ERM}

Under standard ERM, the SAE focuses on minimizing the average loss, which is dominated by Cluster A due to its larger size. This leads to features being optimized primarily to represent Cluster A well. For Cluster B, the reconstruction error is typically higher, leading to less sparse representations (higher $p_B$). This occurs because the network attempts to compensate for poor reconstruction by activating more features, even if they're not ideally suited to the minority cluster's characteristics.

In contrast, Tilted ERM focuses on minimizing the maximum loss, giving more attention to Cluster B. This approach leads to features being adjusted to better represent both clusters. As a result, we expect a slight increase in activation probabilities for Cluster A ($p_A$ increases slightly) as the network makes minor adjustments to accommodate Cluster B. Importantly, we anticipate a significant decrease in activation probabilities for Cluster B ($p_B$ decreases significantly) as the features become more tailored to its characteristics, allowing for sparser and more efficient encoding.

The relationship between feature activation probabilities and reconstruction error is key to understanding the dynamic between ERM and TERM. Lower reconstruction error is associated with lower activation probabilities, as the network can more efficiently encode the input data. Conversely, higher reconstruction error often leads to higher activation probabilities as the network \textit{struggles} to represent the data, activating more features in an attempt to reduce the error.

\subsection{Quantitative Analysis}

To formalize this analysis, let us denote the activation probabilities under ERM as $p_A^{\text{ERM}} = p_A$, $p_B^{\text{ERM}} = p_B$; and under Tilted ERM as $p_A^{\text{Tilted}} = p_A + \Delta p_A$, $p_B^{\text{Tilted}} = p_B - \Delta p_B$.
Here, $\Delta p_A > 0$ is small, reflecting the minor adjustments made to Cluster A's representation, while $\Delta p_B > 0$ is significant, capturing the substantial improvement in Cluster B's encoding.
The difference in total description length is then:
\begin{equation}
\scalebox{0.9}{$
\begin{aligned}
\Delta DL &= DL_{\text{data}}^{\text{ERM}} - DL_{\text{data}}^{\text{Tilted}} \\
&= N_A k \left( H(p_A^{\text{ERM}}) - H(p_A^{\text{Tilted}}) \right) \\
&\quad + N_B k \left( H(p_B^{\text{ERM}}) - H(p_B^{\text{Tilted}}) \right)
\end{aligned}
$}
\end{equation}

\noindent Defining $\Delta H_A = H(p_A^{\text{ERM}}) - H(p_A^{\text{Tilted}})$ and $\Delta H_B = H(p_B^{\text{ERM}}) - H(p_B^{\text{Tilted}})$, we can express this as:
\begin{equation}
\Delta DL = k \left( N_A \Delta H_A + N_B \Delta H_B \right)
\end{equation}
\noindent Our aim is to show that $\Delta DL > 0$ under specific conditions.

\subsection{Conditions for Lower Description Length under Tilted ERM}

For Tilted ERM to yield a lower total description length, we require:
\begin{equation}
N_A \Delta H_A + N_B \Delta H_B > 0
\end{equation}
Given that $N_A \gg N_B$, $\Delta H_A$ is small and negative (since $p_A$ increases slightly), while $\Delta H_B$ is large and positive (since $p_B$ decreases significantly), this condition can be satisfied if:

\begin{equation}
\frac{\Delta H_B}{\left|\Delta H_A\right|} > \frac{N_A}{N_B}
\end{equation}

This inequality summarizes the core of our argument: if the decrease in entropy for Cluster B (per data point) is sufficiently large compared to the increase in entropy for Cluster A, weighted by their respective sample sizes, then Tilted ERM will lead to a lower total description length.

\subsection{Numerical Illustration}

To illustrate this condition, consider a scenario where $p_A^{\text{ERM}} = 0.1$, $p_A^{\text{Tilted}} = 0.12$, $p_B^{\text{ERM}} = 0.5$, and $p_B^{\text{Tilted}} = 0.1$, with $N = 1000$, $N_A = 900$, and $N_B = 100$. Computing the binary entropy values and their differences, we find:
\vspace{-0.05in}
\begin{small}
\begin{align}
\Delta H_A &= H(0.1) - H(0.12) = -0.031 \text{ bits} \nonumber \\
\Delta H_B &= H(0.5) - H(0.1) = 0.531 \text{ bits} \nonumber
\end{align}
\end{small}
\vspace{-0.25in}

\noindent The total difference in description length is then:
\begin{equation}
\scalebox{0.9}{$
\begin{aligned}
\Delta DL &= k \left( 900 \times (-0.031) + 100 \times 0.531 \right) \nonumber \\
&= k \times 25.2 \text{ bits}  \nonumber
\end{aligned}
$}
\end{equation}

\noindent This positive value of $\Delta DL$ demonstrates that, in this example, the total description length is indeed lower under Tilted ERM.

\subsection{Implications}

This proof demonstrates that under specific conditions—namely, when Tilted ERM significantly reduces the activation probabilities for the minority cluster while only slightly increasing them for the majority cluster—the total description length is lower under Tilted ERM compared to standard ERM. According to the MDL principle, which posits that models with lower total description length are preferable, this result implies that Tilted ERM leads to more interpretable features.

The key insight is that Tilted ERM's focus on minimizing the maximum loss allows it to develop features that more efficiently encode both the majority and minority clusters. While this may come at the cost of a slight increase in description length for the majority cluster, the substantial decrease in description length for the minority cluster more than compensates, leading to an overall improvement in feature interpretability.

It's important to note that this analysis relies on several simplifying assumptions, including binary latent codes, independent features, and uniform activation probabilities within clusters. In practice, the actual changes in activation probabilities will depend on the specific data distribution and optimization dynamics. Nonetheless, this theoretical result provides valuable insight into how Tilted ERM can lead to models with enhanced interpretability, particularly in scenarios involving imbalanced datasets.

In future work we could focus on relaxing these assumptions, exploring the implications of continuous-valued latent representations, and investigating the relationship between feature interpretability and other metrics of model performance and fairness. Empirical studies could provide further validation of these theoretical findings across a range of real-world datasets and tasks.

TERM, at high values of the tilt parameter t, can be viewed as minimizing the maximum loss across all data points. Under the assumption that the loss function is proportional to the negative log-likelihood, this becomes equivalent to minimizing the maximum description length in the Minimum Description Length (MDL) framework. In other words, at high tilt, Tilted ERM minimizes the maximum description length for the most poorly represented data points, ensuring that no single data point incurs an excessively long encoding.

This is particularly important in safety-critical applications, such as the detection of rare but hazardous features or circuits. In such cases, these rare features may be infrequent in the dataset and thus underrepresented when training with standard ERM, leading to high description lengths that make detection more difficult. By minimizing the maximum description length through Tilted ERM, these rare safety-relevant features are represented more efficiently, leading to more compact encodings that facilitate their detection and analysis. This improves both the interpretability and reliability of the model, enabling more robust identification of critical features in safety audits or interpretability studies, where compact and clear representations are essential for ensuring that important safety-related circuits are not overlooked.

\section{Applications of Tilted ERM SAEs in Capturing Tail Concepts}
\label{sec:applications}
Sparse Autoencoders trained using ERM focus on minimizing the average reconstruction error across all inputs, leading to strong performance on frequent patterns but poor representation of rare or difficult-to-reconstruct activations. In contrast, SAEs trained via Tilted ERM emphasize reducing the reconstruction error of high-error examples, enabling better capture of rare concepts, improved handling of fine-grained detection tasks, and enhanced performance in high-stakes applications where edge cases are critical. Some applications of TERM-trained SSAEs include:

\subsection{Capturing Tail Concepts in Multilingual Models}

TERM-trained SAEs offer a significant advantage in capturing \emph{rare linguistic patterns}, such as those found in multilingual or dialect-rich datasets. Foundation models (FMs) trained on predominantly English data often struggle to accurately represent less common languages or dialects. While an ERM-trained SAE might prioritize frequent language patterns, a Tilted ERM-trained SAE focuses on reducing reconstruction error for high-error examples, including these rare language patterns.

This approach is particularly important in \emph{multilingual models} used in global applications, where inclusivity and fairness across languages are essential. For example, in a multilingual chatbot, Tilted ERM ensures that features for \emph{low-frequency languages} such as Swahili or Icelandic are reconstructed more accurately, providing a better user experience for speakers of these languages.

\subsection{Fine-Grained Anomaly Detection in High-Stakes Applications}

TERM-trained SAEs excel in \emph{fine-grained anomaly detection}, where identifying subtle deviations from normal behavior is crucial, especially in high-stakes applications such as security, finance, or medical diagnosis. While an ERM-trained SAE might flag anomalies due to their higher-than-average reconstruction error, it is less likely to represent these rare events with sufficient accuracy or disambiguate them from other errors for further analysis.

A Tilted ERM SAE allocates more capacity to rare, high-error cases, ensuring better reconstruction of these outliers. This improves both \emph{detection} and \emph{interpretability} of rare anomalies. For instance, in a financial fraud detection system, a Tilted ERM SAE can better capture the subtle patterns that differentiate fraudulent transactions from normal activity, leading to more effective interventions.

\subsection{Improved Coverage of Rare Concepts in Fairness and Bias Mitigation}

Tilted ERM improves fairness by ensuring that \emph{underrepresented groups} or \emph{tail concepts} are well represented in the model. In many real-world datasets, certain demographic or cultural groups may be underrepresented, leading to \emph{bias in language models}. Tilted ERM ensures that rare patterns—including those associated with underrepresented languages, cultural references, or peoples—are better captured.

This approach leads to a more inclusive model that provides fairer representations across different groups. For example, a Tilted ERM SAE trained on a language model used in customer support applications can better represent \emph{minority dialects} or \emph{regional slang}, reducing bias in customer interactions and improving service for all users.

\subsection{Robustness in Safety-Critical Systems}

In safety-critical systems, such as \emph{autonomous vehicles} or \emph{aviation control}, rare but dangerous events must be handled with high accuracy. Tilted ERM, by focusing on minimizing the reconstruction error for the most difficult cases, ensures that the model is better equipped to handle rare, high-risk scenarios.

For example, in an autonomous driving system, rare but critical inputs such as \emph{uncommon weather conditions} or \emph{unusual road hazards} are more likely to be accurately captured by a Tilted ERM SAE. This improved representation helps the system react more reliably to rare but potentially dangerous situations, enhancing overall safety.

\begin{figure}[htbp]
    \centering
    \includegraphics[width=0.48\textwidth]{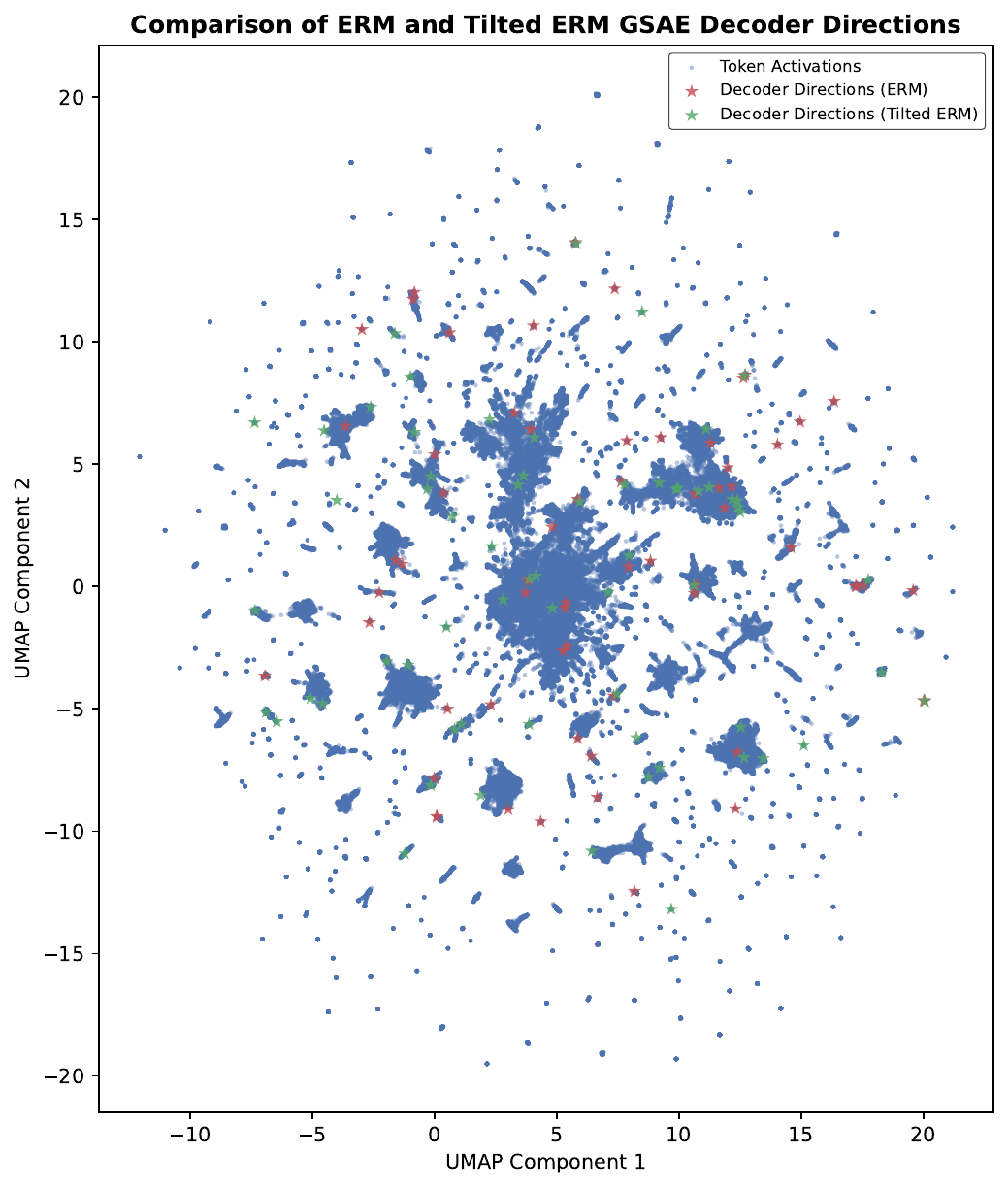}
        \caption{\small UMAP visualization of token activations and decoder features for a TERM-trained and ERM-trained GSAE. Decoder directions for TERM-trained GSAE appear more spread out, suggesting the SAE has wider coverage than the ERM-trained GSAE. }
    \label{fig:umapplot}
\end{figure}

\begin{figure}[htbp]
    \centering
    \includegraphics[width=0.48\textwidth]{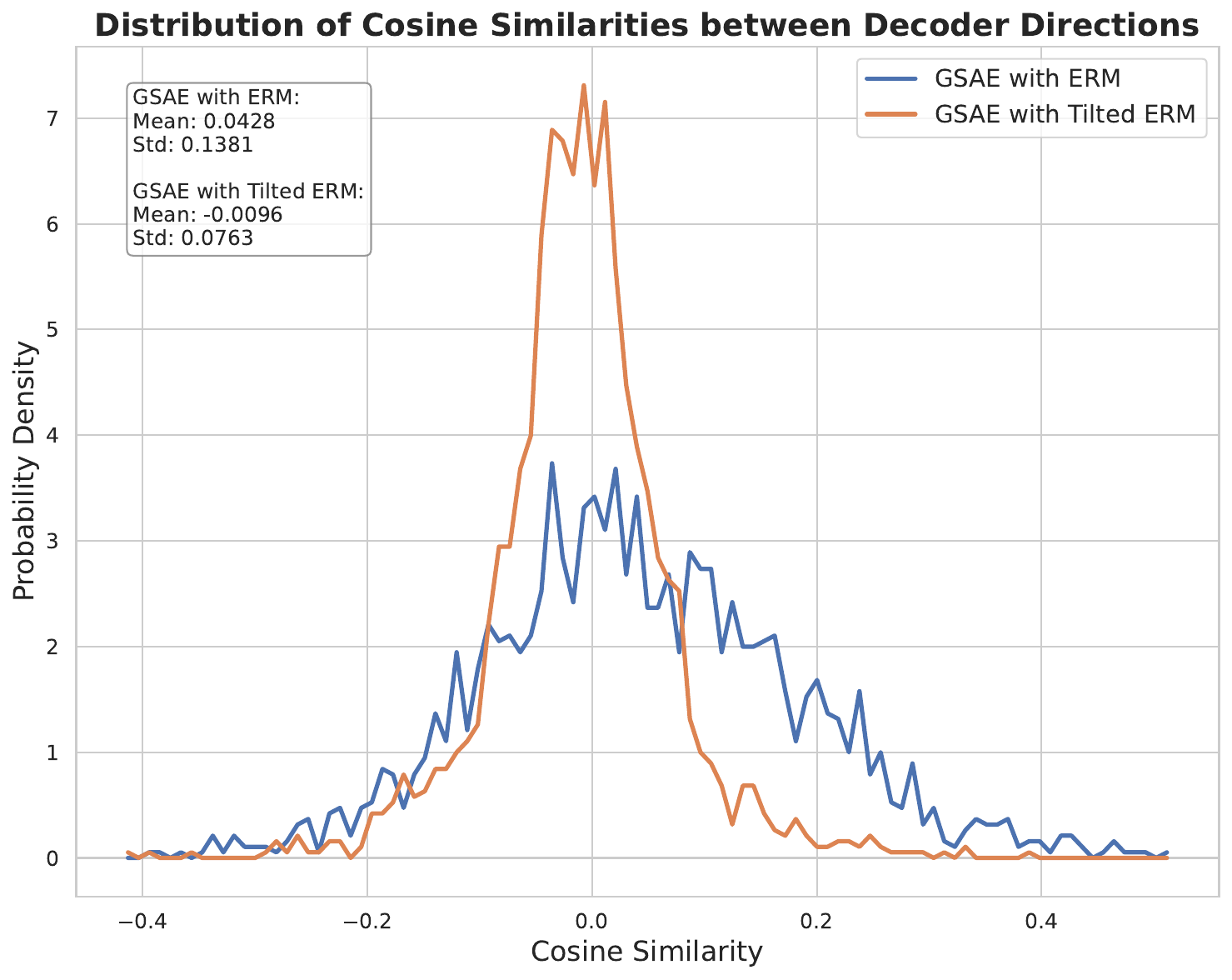}
        \caption{\small Distribution of cosine similarities between decoder directions of TERM-trained and ERM-trained GSAEs. TERM-trained GSAE shows lower similarity between decoder feature directions implying greater coverage. }
    \label{fig:decodercoverage}
\end{figure}

\begin{figure}[htbp]
    \centering
    \includegraphics[width=0.48\textwidth]{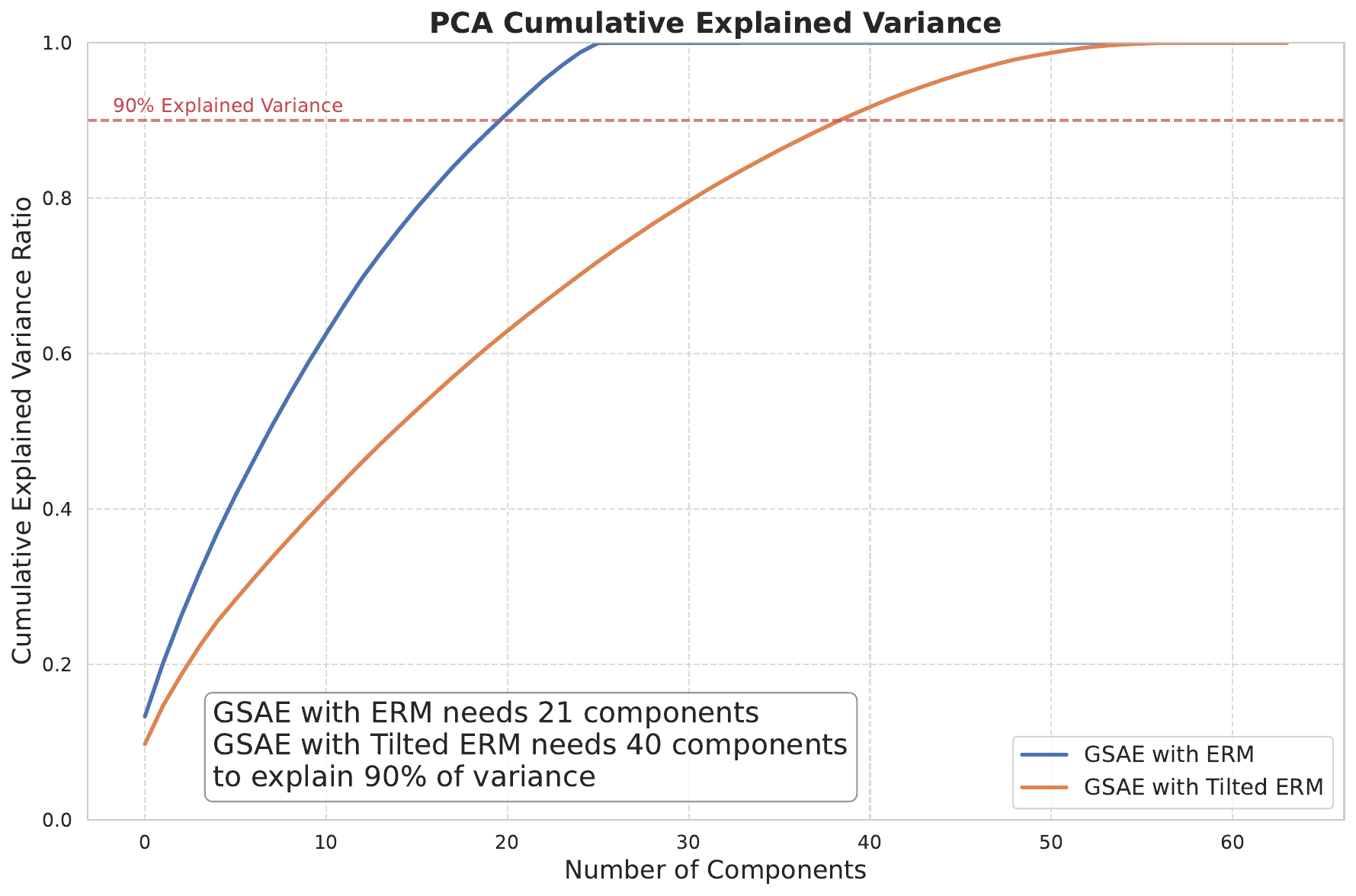}
        \caption{\small \looseness=-1 Number of PCA components required to explain variance in decoder feature directions of TERM-trained and ERM-trained GSAEs. TERM-trained GSAE shows greater variance in decoder feature directions implying greater coverage.}
    \label{fig:pcacoverage}
\end{figure}

\begin{figure}[htbp]
    \centering
    \includegraphics[width=0.48\textwidth]{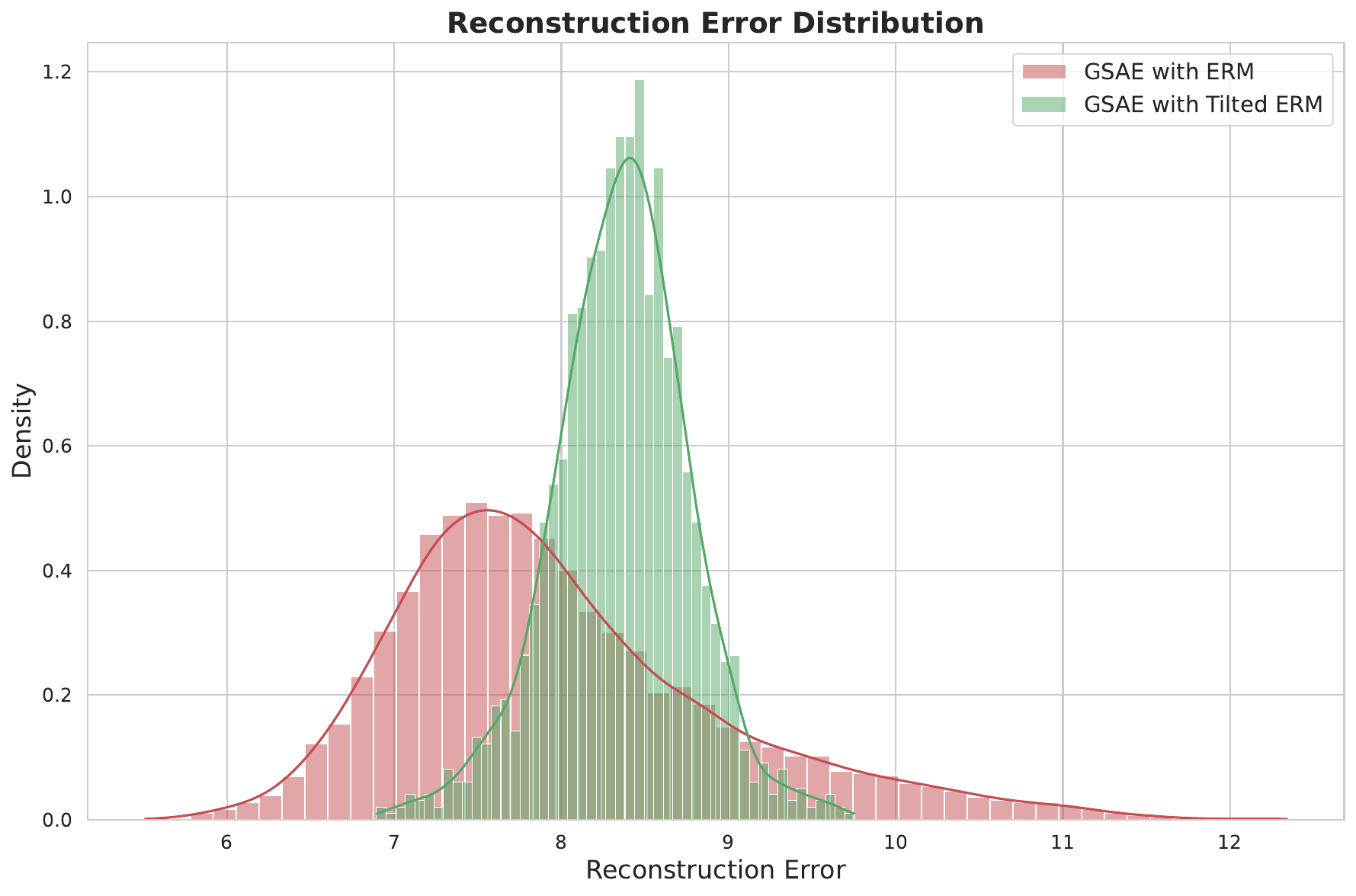}
        \caption{\small Reconstruction error distribution of TERM-trained and ERM-trained GSAE. TERM-trained GSAE minimizes the maximum error at the cost of average error.}
    \label{fig:errorhistogram}
\end{figure}

\begin{figure}[htbp]
    \centering
    \includegraphics[width=0.48\textwidth]{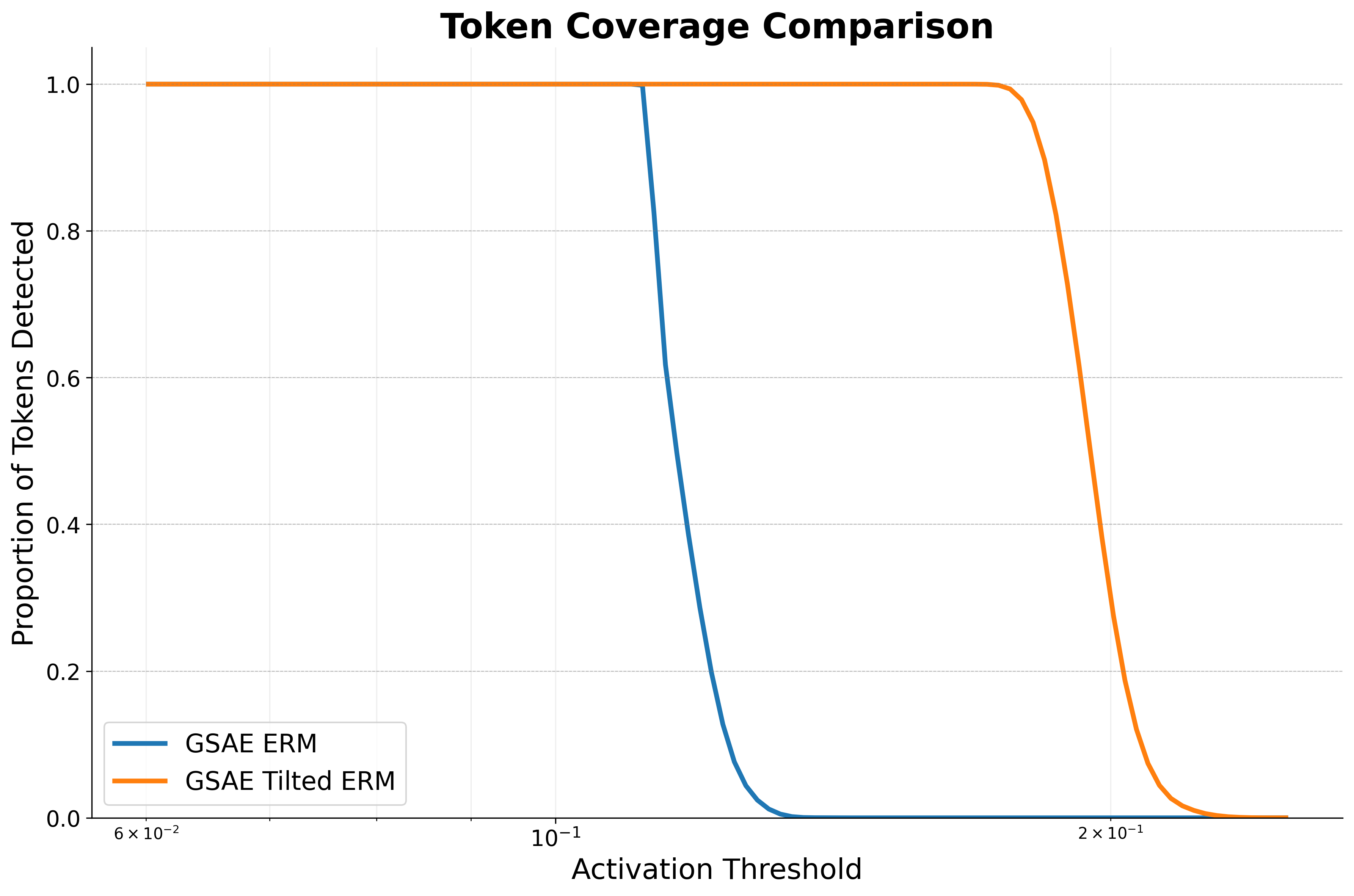}
        \caption{\small Proportion of tokens detected vs. activation threshold for TERM-trained and ERM-trained GSAEs. TERM-trained features exhibit stronger activations.}
    \label{fig:tokensthreshold}
\end{figure}

\section{TERM-trained GSAE Features on TinyStories}

\subsection{UMAP Plot of Decoder Directions}

\autoref{fig:umapplot} plots the UMAP visualization of token activations and decoder features for a TERM-trained and ERM-trained GSAE.

\subsection{Decoder Cosine Similarities}
\autoref{fig:decodercoverage} plots the distribution of cosine similarities between decoder directions of TERM-trained and ERM-trained GSAEs.

\subsection{PCA Components to Explain Variance}

\autoref{fig:pcacoverage} plots the number of PCA components required to explain variance in decoder feature directions of TERM-trained and ERM-trained GSAEs. TERM-trained GSAE shows greater variance in decoder feature directions implying greater coverage.

\subsection{Reconstruction Error}

\autoref{fig:errorhistogram} plots the reconstruction error distribution of TERM-trained and ERM-trained GSAE on 5M tokens sampled from TinyStories. TERM-trained GSAE minimizes the maximum error at the cost of average error.

\subsection{Feature Activation Threshold for TERM-trained and ERM-trained GSAEs}
\autoref{fig:tokensthreshold} plots the proportion of tokens detected vs. activation threshold on 5M tokens from TinyStories for TERM-trained and ERM-trained GSAEs.

\subsection{Feature Diversity}
\label{sec:feature-diversity}
\begin{figure*}[htbp]

\begin{tcolorbox}[colback=gray!5!white,colframe=gray!75!black,title=Box 8: Feature Explanation Aggregation Prompt, width=\textwidth]
\refstepcounter{mytcb}\label{tcb:feature_explanation_aggregation}
\begin{minipage}{\dimexpr\linewidth-2\fboxsep\relax}
\small
\begin{verbatim}
You are an AI assistant tasked with unifying multiple explanations for a single feature in a language
model. These features are from the TinyStories dataset, which consists of short stories using simple 
vocabulary. Your goal is to create a concise explanation that captures the essence of all the
individual explanations.

Individual explanations:
{chr(10).join(f"{i+1}. {exp}" for i, exp in enumerate(explanations))}

Please provide a unified explanation that:
1. Provides a clear and concise description of the feature's function or role in the context of the
TinyStories dataset and the language model. Include 2-3 brief examples of how this feature 
might manifest in the stories.
\end{verbatim}
\end{minipage}
\end{tcolorbox}

\end{figure*}

\begin{figure*}[htbp]
\begin{tcolorbox}[colback=gray!5!white,colframe=gray!75!black,title=Box 9: Diversity Score Generation Prompt, width=\textwidth]
\refstepcounter{mytcb}\label{tcb:diversity_score_generation}
\begin{minipage}{\dimexpr\linewidth-2\fboxsep\relax}
\small
\begin{verbatim}
You are an AI assistant tasked with unifying multiple explanations for a single feature in a language
model. These features are from the TinyStories dataset, which consists of short stories using simple
vocabulary. Your goal is to create a concise explanation that captures the essence of all the 
individual explanations.

Individual explanations:
{chr(10).join(f"{i+1}. {exp}" for i, exp in enumerate(explanations))}

Please provide a unified explanation that:
1. Provides a clear and concise description of the feature's function or role in the context of the 
TinyStories dataset and the language model. Include 2-3 brief examples of how this feature might
manifest in the stories.
2. Scores the diversity of the feature's activations on a scale of 1 to 100, where:
   - 1-20: Very low diversity (e.g., a specific feature that only activates for a specific character
   name like "Tom")
   - 21-40: Low diversity (e.g., a less generic feature that activates for different character names, 
   but only names)
   - 41-60: Moderate diversity (e.g., a generic feature that activates for various types of objects 
   found in a home)
   - 61-80: High diversity (e.g., a generic feature that activates for different types of actions, both
   physical and verbal)
   - 81-100: Very high diversity (e.g., a generic feature that activates across various story elements: 
   characters, actions, settings, emotions, dialogue)
Note: Consider the full range of possibilities within the TinyStories dataset. Don't hesitate to use
the full scale from 1 to 100 based on your analysis even if they all pertain to children's stories
since this is the dataset we are evaluating.

Unified explanation:
[Your unified explanation with 2-3 examples]

Diversity Score: [1-100]
Justification:[Brief justification for the score, considering the context of the TinyStories dataset]
\end{verbatim}
\end{minipage}
\end{tcolorbox}

\end{figure*}
When comparing the feature diversity score distribution of TERM-trained GSAE with ERM-trained GSAE in \autoref{fig:gsaedist}, we observe that TERM-trained GSAE induces some features to specialize in tail concepts, while others generalize to represent a broader range of concepts relative to the ERM-trained GSAE.

To generate this plot, we first extract explanations for features based on the input examples that activate them, using the prompt detailed in Box \ref{tcb:interpreter_prompt}. To understand the behavior of all features, particularly those representing tail concepts, we cannot use random sampling of the TinyStories dataset, as employed in prior work since this would not capture tail concepts effectively. Instead, we process the entire TinyStories dataset in chunks of 5 million data points, and generate explanations by sampling uniformly from the top 50\% examples that activate a feature. We then aggregate the explanations from each chunk using the explanation aggregation prompt provided in Box \ref{tcb:feature_explanation_aggregation}.

After aggregating feature explanations across dataset chunks, we derive the diversity score. This score is obtained using the score generation prompt presented in Box \ref{tcb:diversity_score_generation}, implemented with Claude 3.5 Sonnet (\texttt{claude-3-5-sonnet-20240620}).

\section{Dataset Details}
\label{sec:datasetdetails}
All datasets used in this study are in English. Below are the details for each dataset:

\paragraph{OpenWebText (OWT)} A large-scale, diverse corpus of web content derived from URLs shared on Reddit. We use a single split comprising approximately 8 million documents and over 40GB of text data.\footnote{\url{https://huggingface.co/datasets/Skylion007/openwebtext}}
\paragraph{Pile} We utilize 2B tokens from the Pile dataset, a large-scale curated corpus designed for language model training. This subset contains 10.8M examples across various domains including academic writing, code, and web content.\footnote{\url{https://huggingface.co/datasets/NeelNanda/pile-small-tokenized-2b}}
\paragraph{TinyStories} A dataset of simple, coherent stories generated specifically for language model research. It consists of a single split containing 2.12M training examples, designed to be semantically meaningful while using limited vocabulary.\footnote{\url{https://huggingface.co/datasets/roneneldan/TinyStories}}
\paragraph{arXivPhysics} A collection of physics papers from arXiv. We use the first five examples, comprising 4.8M tokens. The full dataset contains 15.8k rows, split into 60\% train, 20\% validation, and 20\% test, representing a broad range of physics topics.\footnote{\url{https://huggingface.co/datasets/anonymousdatasets/arxiv-physics}}
\paragraph{Physics Instruction Tuning} A specialized dataset for physics-related instruction tuning. We use all 700K tokens from this dataset, which contains 30k examples of physics questions, explanations, and problem-solving instructions.\footnote{\url{https://huggingface.co/datasets/AlgorithmicResearchGroup/arxiv-physics-instruct-tune-30k}}
\paragraph{Pile Toxicity} A curated subset of the Pile dataset focusing on toxic content, designed for studying and mitigating harmful language in language models. We employ a 60-20-20 train-validation-test split to ensure balanced evaluation.\footnote{\url{https://huggingface.co/datasets/tomekkorbak/pile-toxicity-balanced}}
\paragraph{Bias in Bios} A dataset of online biographies used to study gender bias in machine learning models. It contains 257k training examples, 39.6k validation examples, and 99.1k test examples, providing a rich source for analyzing gender representation in professional contexts.\footnote{\url{https://huggingface.co/datasets/LabHC/bias_in_bios}}

\section{Computational Resources}

Our experiments were conducted using modest computational resources, showing the accessibility of our approach. All experiments, including:
\begin{itemize} [itemsep=0pt, topsep=0pt, leftmargin=10pt, rightmargin=1pt, labelsep=2pt, itemindent=2pt, parsep=0pt]
\item Finetuning SSAEs on OpenWebText, Physics-arXiv, Toxicity data, and Pile datasets
\item Training the Pythia-70M classifier and other baselines for the Bias in Bios task
\item Pretraining GSAEs on the TinyStories dataset
\end{itemize}
\noindent were completed using 4 NVIDIA A100 GPUs or A6000 GPUs in less than 24 hours.

\section{TERM-trained and ERM-trained GSAE features on TinyStories}
\label{sec:tinystories-features}

We present a qualitative analysis of the feature explanations derived from both TERM-trained and ERM-trained GSAEs on the TinyStories dataset. Four explanations for each SSAE are shown in Tables \ref{tab:ermgsae} and \ref{tab:termgsae}.

\paragraph{TERM-trained GSAE}
TERM-trained GSAEs exhibit a fascinating mix of features, some capturing broad conceptual themes while others specialize in highly specific linguistic patterns. This duality stems from TERM's objective of minimizing the maximum loss, encouraging the SAE to learn features that can effectively reconstruct both frequent and rare examples.

Feature \texttt{h.7\_feature17}, for example, is remarkably broad, described as processing ``text related to children's stories, simple narratives, and basic concepts in children's literature.'' This wide scope allows it to represent various story elements, from character actions and emotions to dialogue and sensory experiences, reflecting TERM's focus on capturing the full spectrum of data patterns.

In contrast, feature \texttt{h.7\_feature8} demonstrates TERM's ability to learn highly specific features. It activates exclusively on the indefinite article ``an'' when introducing new elements in a story, suggesting its role in recognizing a distinct grammatical pattern within the TinyStories dataset. This specific feature might capture a unique characteristic of the data or potentially represent a less frequent but important narrative element.

\paragraph{ERM-trained GSAE}
ERM-trained GSAE features, on the other hand, tend towards greater specificity, reflecting ERM's focus on minimizing the average reconstruction error. This leads to features that accurately represent the most common patterns in the data but might struggle to capture tail concepts effectively.

For instance, feature \texttt{h.7\_feature3 }in the ERM-trained GSAE is tailored to recognizing ``narrative structures in simple, moralistic children's stories.'' While it encompasses a range of story elements, its scope remains constrained to a specific type of narrative common within the TinyStories dataset. This contrasts with the broader TERM feature \texttt{h.7\_feature17}, which captures the essence of children's stories more generally.

\paragraph{Implications}
This qualitative analysis suggests that TERM, by balancing broad and specific features, encourages the learning of more compositional representations, potentially improving the SAE's ability to detect and interpret a wider variety of concepts, including rare or underrepresented ones. ERM's emphasis on specificity, while effective for frequent patterns, may limit the SAE's capacity to accurately represent the full spectrum of data patterns, particularly those found in the tail of the distribution.

\begin{featureset}{ERM-trained GSAE Features}
\label{tab:ermgsae}
\featurerow{h.7\_feature3}{%
Unified explanation: This neuron recognizes narrative structures in simple, moralistic children's stories. It activates on new story segments, character introductions, settings, conflicts, and dialogue. Frequent themes include lessons on kindness, honesty, and sharing.

Examples:
\begin{enumerate}[leftmargin=*, nosep]
  \item "Lily woke up early on Saturday morning. 'Mom, can I go play with my friend Jenny?' she asked."
  \item "Once upon a time, there was a little boy named Tommy who loved to play with his toys but never wanted to share."
  \item "After school, Timmy came home feeling sad. 'What's wrong?' his mom asked. 'I got in trouble for not telling the truth,' Timmy replied."
\end{enumerate}

Diversity Score: 71

Justification: Activates on diverse narrative elements in children's stories, including dialogue, character introductions, settings, events, emotions, and moral lessons. High diversity within the genre of educational stories for young audiences. \\
}

\featurerow{h.7\_feature5}{%
Unified explanation: This neuron activates on language patterns associated with conveying moral lessons, advice, and guidance on appropriate behavior in children's stories or parental scenarios. It frequently fires on modal verbs like "should" and "can" when characters are learning about right and wrong actions, facing consequences, or being instructed on proper conduct.

Examples:
\begin{enumerate}[leftmargin=*, nosep]
  \item "You should not take things that don't belong to you," said Mom, after catching Timmy taking a candy bar from the store.
  \item "The little boy learned that he can be kind to others by sharing his toys."
  \item "If you can't say something nice, you should not say anything at all," advised the teacher to the rowdy class.
\end{enumerate}

Diversity Score: 68

Justification: While specializing in moral lessons and guidance, the range of potential lessons, advice, and behavioral instructions is quite broad. It activates across various story elements and moral themes, encompassing a diverse array of instructional language in children's literature. \\
}

\featurerow{h.7\_feature6}{%
Unified Explanation: This neuron activates when "<|endoftext|>" is followed by the beginning of a short, simple story or narrative, often with a moral lesson, cautionary tale, or tragic ending. These stories frequently feature children or animals as main characters, written in a style suitable for young readers.

Examples:
\begin{enumerate}[leftmargin=*, nosep]
  \item "<|endoftext|> Once upon a time, there was a little girl who loved to play in the forest. One day, she wandered too far from home and got lost..."
  \item "<|endoftext|> A group of young animals decided to explore the old abandoned barn, despite their parents' warnings. But it was too late when they realized the danger inside..."
  \item "<|endoftext|> Tommy was a curious boy who couldn't resist the temptation of the old well in his backyard. He leaned over too far and..."
\end{enumerate}

Diversity Score: 61

Justification: While specific to children's stories, the diversity is high, involving various characters, settings, actions, and themes. It captures a range of narrative elements, including plot structure, character archetypes, and common literary devices. \\

}

\featurerow{h.7\_feature12}{%
Unified explanation: This neuron activates at the beginning of short stories or narratives aimed at children. The consistent trigger is the token "<|endoftext|>", indicating the start of a new text sample. It recognizes the opening of simple narrative structures, often involving young protagonists, animal characters, moral lessons, or elements of danger or misfortune.

Examples:
\begin{enumerate}[leftmargin=*, nosep]
  \item "<|endoftext|> Once upon a time, there was a little girl named Lily who loved to explore the enchanted forest near her home."
  \item "<|endoftext|> In a cozy burrow, a family of rabbits lived happily until a hungry fox threatened their safety."
  \item "<|endoftext|> Tommy the turtle was always in a hurry, but his impatience nearly cost him his life when he wandered too far from home."
\end{enumerate}

Diversity Score: 71

Justification: While focused on children's stories, the range of possible stories and themes is quite diverse, involving different characters, settings, plots, and outcomes.
}

\end{featureset}

\begin{featureset}{TERM-trained GSAE Features}
\label{tab:termgsae}
\featurerow{h.7\_feature8}{%
Unified explanation: This feature detects the indefinite article "an" when introducing new or significant elements in children's stories or simple narratives. It activates when "an" precedes a noun at the beginning of a sentence or clause, signaling a novel element important to the plot.

Examples:
\begin{enumerate}[leftmargin=*, nosep]
  \item "An old man lived in a tiny house by the forest."
  \item "One day, an unexpected visitor arrived at the village."
  \item "Deep in the ocean, an ancient treasure awaited discovery."
\end{enumerate}

Diversity Score: 65

Justification: High diversity in types of elements introduced (characters, objects, concepts) within children's stories, but limited to narrative contexts. \\
}

\featurerow{h.7\_feature13}{%
Unified explanation: This feature captures interjections or exclamations in children's stories or dialogues expressing surprise, excitement, or drawing attention to something noteworthy. Tokens like "Wow" or "Look" often appear at the beginning of quoted speech or exclamations.

Examples:
\begin{enumerate}[leftmargin=*, nosep]
  \item "Wow! Look at that giant castle!" a child might exclaim upon seeing an impressive structure.
  \item "Look, the caterpillar turned into a butterfly!" a character might say, pointing out a transformation.
  \item "Wow, that was a close one!" someone might remark after narrowly avoiding danger.
\end{enumerate}

Diversity Score: 71

Justification: While specific to interjections, these can be used across a wide range of contexts and story elements, reflecting a high degree of diversity within children's stories and dialogues. \\
}

\featurerow{h.7\_feature14}{%
Unified explanation: This neuron predicts words related to pleasant or appetizing food experiences in children's stories or simple narratives. It activates on the first few letters of words like "yummy", "candy", "crumbs", and "celery", generating vocabulary associated with tasty treats, cooking, or domestic activities.

Examples:
\begin{enumerate}[leftmargin=*, nosep]
  \item "The little girl licked her lips as she stared at the yummy chocolate cake."
  \item "After playing outside, the kids ran to the kitchen for a snack of celery and peanut butter."
  \item "Mom swept up the crumbs from the cookies the children had enjoyed earlier."
\end{enumerate}

Diversity Score: 53

Justification: While primarily focused on food-related words, it recognizes a range of vocabulary including adjectives, nouns, and verbs related to food experiences in children's stories. \\
}

\featurerow{h.7\_feature17}{%
Unified explanation: This neuron processes text related to children's stories, simple narratives, and basic concepts in children's literature. It responds to character names, diminutives, dialogue markers, sensory experiences, emotions, onomatopoeias, common objects, food items, childhood experiences, simple actions, and basic vocabulary.

Examples:
\begin{enumerate}[leftmargin=*, nosep]
  \item "Ducky waddled over to the lollipop on the ground. 'Yum!' he exclaimed, gobbling it up."
  \item "Ow, ow, ow! Timmy had scraped his knee on the rough sand. Mom kissed it better and gave him a sausage to cheer him up."
  \item "Bark, bark! Spidey's new puppy was digging in the garden, scattering the soil everywhere. 'No, no, pup!' scolded Spidey."
\end{enumerate}

Diversity Score: 85

Justification: Displays very high diversity within children's literature, responding to a wide range of elements including characters, emotions, actions, objects, sensory experiences, and dialogue patterns.
}

\end{featureset}

\end{document}